\documentclass[fleqn,10pt]{wlscirep}
\usepackage[utf8]{inputenc}
\usepackage[T1]{fontenc}

\usepackage[bottom]{footmisc}
\usepackage{fnpct}
\usepackage{afterpage}
\usepackage{xcolor}
\usepackage{hyperref}
\usepackage{tabularx}
\usepackage{booktabs}
\newcolumntype{P}[1]{>{\centering\arraybackslash}p{#1}}
\newcolumntype{C}{>{\Centering}X}    
\newcolumntype{L}{>{\RaggedRight}X}  
\usepackage{algorithm}  
\usepackage{algorithmic}
\usepackage{multirow}
\usepackage{graphicx}
\usepackage{enumitem}
\usepackage{caption, subcaption}
\usepackage{cleveref}
\usepackage{threeparttable}
\usepackage{comment}
\usepackage{graphicx}
\usepackage{multirow}
\usepackage{longtable}
\usepackage{graphicx}
\expandafter\def\expandafter\normalsize\expandafter{%
    \normalsize
    \setlength\abovedisplayskip{2pt}
    \setlength\belowdisplayskip{2pt}
    \setlength\abovedisplayshortskip{2pt}
    \setlength\belowdisplayshortskip{2pt}
}

\RequirePackage{amsfonts}

\newcommand{\cblue}{\color{blue}}

\newcolumntype{L}[1]{>{\raggedright\arraybackslash}p{#1}}
\newcolumntype{C}[1]{>{\centering\arraybackslash}p{#1}}
\newcolumntype{R}[1]{>{\raggedleft\arraybackslash}p{#1}}

\usepackage[resetlabels]{multibib}
\newcites{ec}{References}

\newcites{onerevse}{References}
\newcites{onerevone}{References}
\newcites{onerevtwo}{References}
\newcites{onerevthree}{References}

\usepackage[toc,page]{appendix}

\title{Bias of AI-Generated Content: An Examination of News Produced by Large Language Models}

\author[1,*]{Xiao Fang}
\author[2,*]{Shangkun Che}
\author[1,+]{Minjia Mao}
\author[3,+]{Hongzhe Zhang}
\author[1,+]{Ming Zhao}
\author[4,+]{Xiaohang Zhao}
\affil[1]{University of Delaware, U.S.A.}
\affil[2]{Tsinghua University, China}
\affil[3]{Chinese University of Hong Kong, Shenzhen, China}
\affil[4]{Shanghai University of Finance and Economics, China}

\affil[*]{corresponding authors: xfang@udel.edu, csk19@mails.tsinghua.edu.cn}

\affil[+]{these authors are listed in alphabetical order}

\begin{abstract}
Large language models (LLMs) have the potential to transform every aspect of our lives and work through the content they generate, known as AI-Generated Content (AIGC). To harness the advantages of this transformation, it is essential to understand the limitations of LLMs. Here, we investigate the bias of AIGC produced by seven representative LLMs, encompassing early models like Grover and recent ones such as ChatGPT, Cohere, and LLaMA. In our investigation, we collect 8,629 news articles from The New York Times and Reuters, both of which are highly ranked in terms of their dedication to provide accurate and unbiased news, over the period of December 2022 to April 2023. We then apply each examined LLM to generate news content with headlines of these news articles as prompts, and evaluate the gender and racial biases of the AIGC produced by the LLM by comparing the AIGC and the original news articles at the word, sentence, and document levels. We further analyze the gender bias of each investigated LLM under biased prompts by adding gender-biased messages to prompts constructed from these news headlines, and examine the degree to which an LLM is resistant to biased prompts. 

\vspace{0.2cm}

\hspace{0.2cm} Our study reveals that the AIGC produced by each examined LLM deviates substantially from the news articles collected from The New York Times and Reuters, in terms of word choices related to gender or race, expressed sentiments and toxicities towards various gender or race-related population groups in sentences, and conveyed semantics concerning various gender or race-related population groups in documents. Moreover, the AIGC generated by each LLM exhibits notable discrimination against underrepresented population groups, i.e., females and individuals of the Black race. Among the investigated LLMs, the AIGC generated by ChatGPT demonstrates the lowest level of bias, which is partly attributed to its reinforcement learning from human feedback (RLHF) feature. Furthermore, thanks to its RLHF feature, ChatGPT is the sole model capable of declining content generation when provided with biased prompts. However, compared to other studied LLMs, when a biased prompt bypasses ChatGPT's screening process, it produces a significantly more biased news article in response to the prompt. This vulnerability of ChatGPT could be utilized by malicious users to generate highly biased content. 

\end{abstract}

\begin{document}

\flushbottom
\maketitle
%
%
\thispagestyle{empty}


\section*{Introduction}

Large language models (LLMs), such as ChatGPT and LLaMA, are large-scale AI models trained on massive amounts of data to understand human languages \cite{ouyang2022training,touvron2023llama}. Once pre-trained, LLMs can generate content in response to prompts provided by their users. Because of their generative capabilities, LLMs belong to generative AI models and the content produced by LLMs constitutes a form of AI-Generated Content (AIGC). Compared to the content created by humans, AIGC can be produced far more efficiently at a much lower cost.
As a result, LLMs have the potential to facilitate and revolutionize various kinds of work in organizations: from generating property descriptions for real estate agents to creating synthetic patient data for drug discovery \cite{StandfordHAI2013}. To realize the full benefit of LLMs, we need to understand their limitations \cite{StandfordHAI2013}. LLMs are trained on archival data produced by humans. Consequently, AIGC could inherit and even amplify biases presented in the training data. Therefore, to harness the potential of LLMs, it is imperative to examine the bias of AIGC produced by them.

\begin{figure}[!b]
\begin{center}
	\includegraphics[scale=0.2]{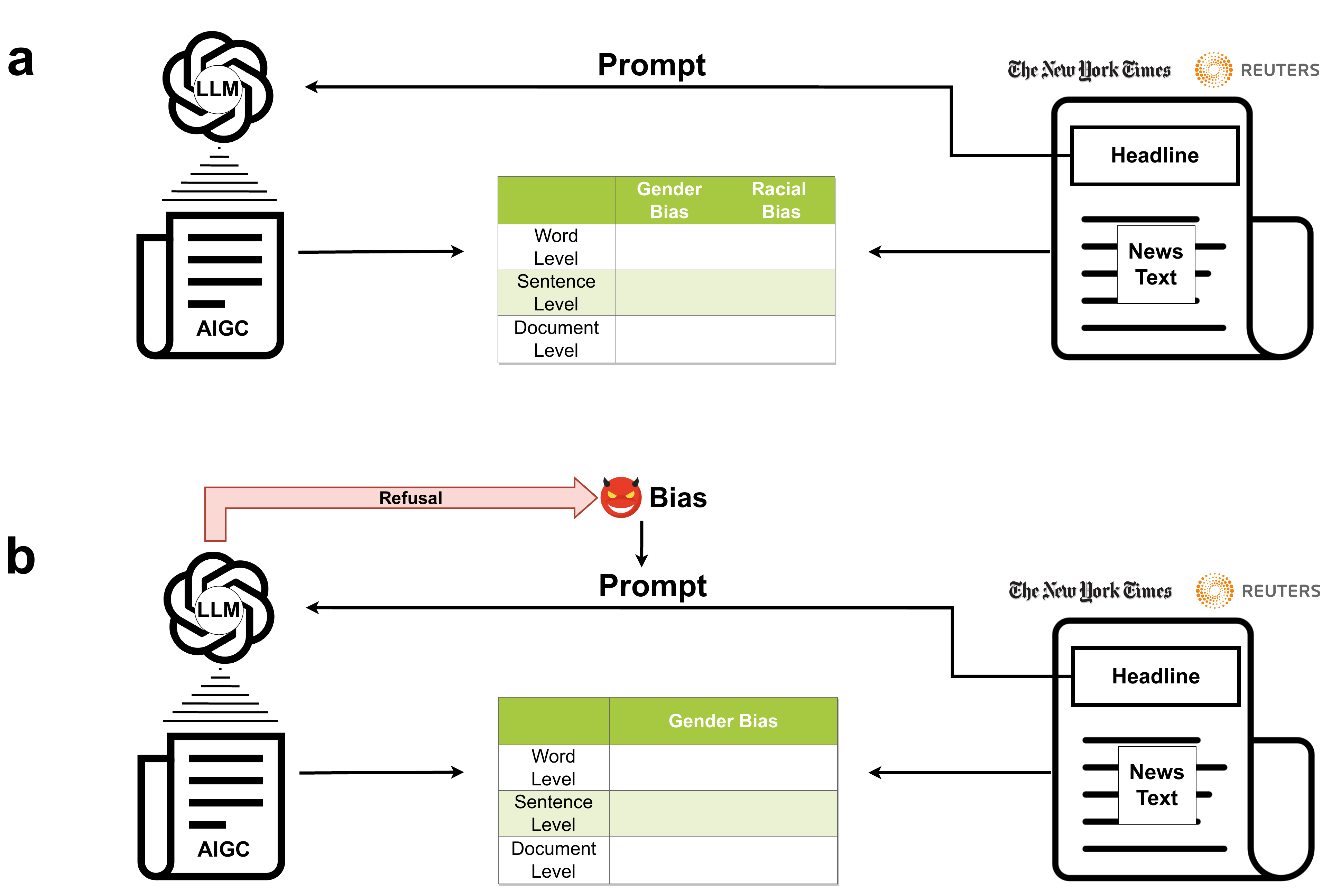}
\end{center}
\caption{\textbf{Framework for Evaluating Bias of AIGC.} \textbf{a} We proxy unbiased content  with the news articles collected from The New York Times and Reuters. Please see the Subsection of ``Data'' for the justification of choosing these news agencies. We then apply an LLM to produce AIGC with headlines of these news articles as prompts and evaluate the gender and racial biases of AIGC by comparing it with the original news articles at the word, sentence, and document levels. \textbf{b} Examine the gender bias of AIGC under biased prompts.}
\label{fg:framework}
\end{figure}

In general, bias refers to the phenomenon that  computer systems ``systematically and unfairly discriminate against certain individuals or groups of individuals in favor of others'' \cite[p.332]{friedman1996bias}. 
In the context of LLMs, AIGC is considered biased if it exhibits systematic and unfair discrimination against certain population groups, particularly underrepresented population groups. 
Among various types of biases, gender and racial biases have garnered substantial attention across multiple disciplines 
\cite{guglielmi2018gender,obermeyer2019dissecting, centola2021reduction, baker2021algorithmic, galos2023gender}, 
due to their profound impacts on individuals and wide-ranging societal implications. 
Accordingly, we investigate the gender and racial biases of AIGC by examining their manifestation in AIGC at the word, sentence, and document levels. At the word level, 
the bias of AIGC is measured as the degree to which the distribution of words associated with different population groups (e.g., male and female) in AIGC deviates from a reference distribution \cite{beukeboom2019stereotypes,liang2022holistic}. Clearly, the choice of the reference distribution is critical to the bias evaluation. Previous studies utilize the uniform distribution as the reference distribution \cite{liang2022holistic}, which might not reflect the reality of content production. For example, a news article about the Argentina men's soccer team winning the 2022 World Cup likely contains more male-specific words than female-specific words, thereby deviating from the uniform distribution. The reason for this deviation may be attributed to the fact that the news is about men's sports and male players, rather than favoring one population group over another. Ideally, we would derive a reference distribution from unbiased content. However, in reality, completely unbiased content may not exist. 
One viable solution is to proxy unbiased content with news articles from news agencies that are highly ranked in terms of their dedication to provide accurate and unbiased news, and obtain a reference distribution from these news articles. 
To evaluate the word-level bias of AIGC created by an LLM, we apply the LLM to generate news content with headlines of these news articles as prompts. We then compare the distribution of words associated with different population groups in the generated content to the reference distribution. 
It is noted that LLMs have been utilized to generate news. For example, a recent work employs headlines from the New York Times as prompts for producing LLM-generated news \cite{munoz-ortiz_contrasting_2023_article}. Additionally, the use of LLMs to write the first draft of articles is reported in the media \cite{davenport_how_2022_article}. 

At the sentence level, the bias of AIGC is evaluated as the difference between sentences related to each investigated population group in AIGC and their counterparts in the news articles produced by the selected news agencies, in terms of their expressed sentiments and toxicities. In this study, we define a document as a news article that is either produced by a news agency or generated by a LLM. And the document-level bias of AIGC is assessed as the difference between documents in AIGC and their counterparts produced by the selected news agencies, in terms of their expressed semantics regarding each investigated population group. 
These three levels of evaluations vary in their unit of analysis, spanning from words to sentences and documents, and examine the bias of AIGC from different aspects of content generation. Specifically, the word-level analysis uncovers the bias of AIGC in word choices, the sentence-level investigation reveals its bias in assembling words to express sentiments and toxicities, while the document-level examination discloses its bias in organizing words and sentences to convey the meaning and themes of a document.
Moreover, a malicious user could intentionally supply biased prompts to an LLM and induce it to generate biased content. Therefore, it is necessary to analyze the bias of AIGC under biased prompts. To this end, we add biased messages to prompts constructed from news headlines and examine the gender bias of AIGC produced by an LLM under biased prompts.
In addition, a well-performed LLM could refuse to generate content when presented with biased prompts. Accordingly, we also assess the degree to which an LLM is resistant to biased prompts. 
Figure~\ref{fg:framework} depicts the framework for evaluating the bias of AIGC.

It is important to note that bias is not a unique challenge limited to AIGC. Within journalism research \cite{leppanen2020automated}, bias is a prevalent issue observed in news articles. An illustrative example is when both a man and a woman contribute equally to an event, yet prior research reveals a tendency in news reporting to place the man's contribution first. This position bias results in greater visibility for the man. Other studies delve into association bias between different genders and specific words in language generation \cite{sheng2019woman}. Despite numerous efforts proposing debiasing methods, researchers argue \cite{gonen2019lipstick} that these approaches merely mask bias rather than truly eliminating it from the embedding. As language models grow in size, they become more susceptible to acquiring biases against marginalized populations from internet texts \cite{bender2021dangers}. Our evaluation differs from existing studies that primarily focus on evaluating the bias manifested in \textit{short phrases} (e.g., a word or a sentence) generated by language models \cite{huang2019reducing, nadeem2021stereoset, liang2021towards, kirk2021bias}. For example, Nadeem et al. (2021) develop the context association test to evaluate the bias of language models, which essentially asks an language model to choose an appropriate word or sentence in a given context (e.g. Men tend to be more [soft or determined] than women) \cite{nadeem2021stereoset}. Our study, on the other hand, evaluates the bias embodied in \textit{documents} (i.e., news articles) produced by an LLM, in terms of its word choices as well as expressed sentiments, toxicities, and semantics in these documents.

\section*{Results}

\subsection*{Word Level Bias}
 
\subsubsection*{Gender Bias}

\begin{figure}[!b]
 \centering
	\includegraphics[scale=0.9]{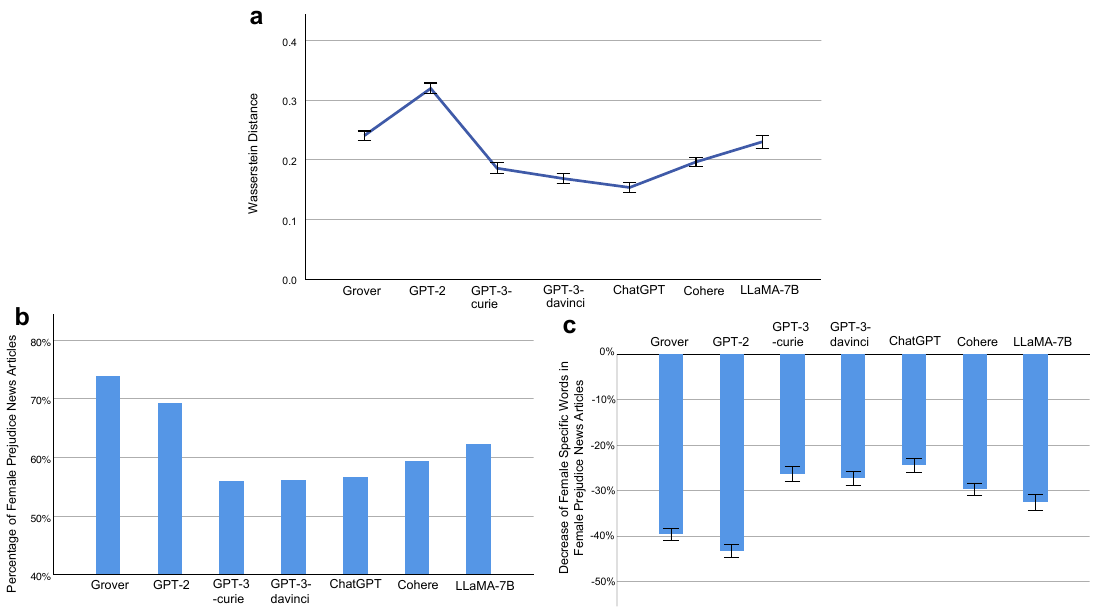}
\caption{\textbf{Gender Bias at Word Level.} \textbf{a} Word level gender bias of an LLM and its 95\% confidence interval (error bar), measured using the average Wasserstein distance defined by Equation~\ref{eq:W_bar}. For example, the gender bias score of 0.2407 by Grover indicates that, on average, the absolute difference between the percentage of male (or female) specific words out of all gender related words in a news article generated by Grover and that percentage in its counterpart collected from The New York Times or Reuters is 24.07\%. \textbf{b} Percentage of female prejudice news
articles generated by an LLM. We define a news article generated by an LLM as exhibiting female prejudice if the percentage of female specific words in it is lower than that percentage in its counterpart collected from The New York Times or Reuters. \textbf{c} Decrease of female specific words in female prejudice news articles
generated by an LLM and its 95\% confidence interval (error bar). For example, the score of -39.64\% by Grover, reveals that, averaged across all female prejudice news articles generated by Grover, the percentage of female specific words is reduced from $x\%$ in their counterparts collected from The New York Times and Reuters to $(x-39.64)\%$ in those generated by Grover.   }
\label{fg: word_gender}
 
\end{figure}

Measured using the average Wasserstein distance defined by Equation~\ref{eq:W_bar}, the word level gender biases of the examined LLMs are: Grover 0.2407 (95\% CI [0.2329, 0.2485], $N=5,671$), GPT-2 0.3201 (95\% CI [0.3110, 0.3291], $N=5,150$), GPT-3-curie 0.1860 (95\% CI [0.1765, 0.1954], $N=3,316$), GPT-3-davinci 0.1686 (95\% CI [0.1604, 0.1768], $N=3,807$), ChatGPT 0.1536 (95\% CI [0.1455, 0.1617], $N=3,594$), Cohere 0.1965 (95\% CI [0.1890, 0.2041], $N=5,067$), and LLaMA-7B 0.2304 (95\% CI [0.2195, 0.2412], $N=2,818$) (Figure~\ref{fg: word_gender}a). Overall, the AIGC generated by each investigated LLM exhibits substantial word level gender bias. Among them, ChatGPT achieves the lowest score of gender bias. Specifically, its score of 0.1536 indicates that, on average, the absolute difference between the percentage of male (or female) specific words out of all gender related words in a news article generated by ChatGPT and that percentage in its counterpart collected from The New York Times or Reuters is 15.36\%. Notably, among the four GPT models, the word level gender bias decreases as the model size increases from 774M (GPT-2) to 175B (GPT-3-davinci, ChatGPT). Moreover, although GPT-3-davinci and ChatGPT have the same model size, the outperformance of ChatGPT over GPT-3-davinci highlights the effectiveness of its RLHF (reinforcement learning from human feedback) feature in mitigating gender bias. 


Figure~\ref{fg: word_gender}a presents the overall gender bias of each investigated LLM, which reveals its magnitude of deviation from the reference human-writing in terms of its utilization of gender related words. 
It is interesting to zoom in and uncover the bias of each LLM against the underrepresented group (i.e., females in the focal analysis).
To this end, we define a news article generated by an LLM as showing female prejudice if the percentage of female specific words in it is lower than that percentage in its counterpart collected from The New York Times or Reuters. 
Figure~\ref{fg: word_gender}b reports the proportion of female prejudice news articles generated by each LLM : Grover 73.89\% ($N=3,037$), GPT-2 69.24\% ($N=2,828$), GPT-3-curie 56.04\% ($N=1,954$), GPT-3-davinci 56.12\% ($N=2,183$), ChatGPT 56.63\% ($N=2,045$), Cohere 59.36\% ($N=2,879$), and LLaMA-7B 62.26\% ($N=1,627$). Let us consider Grover's performance of 73.89\% as an example. This figure suggests that, for a news article obtained from the New York Times or Reuters that includes female specific words, there is a probability of 0.7389 that the percentage of female specific words in its corresponding news article generated by Grover is lower than that percentage found in the original article.
Moreover, Figure~\ref{fg: word_gender}c shows the extent of the decrease in female specific words in female prejudice news articles generated by each investigated LLM. As shown, such measurements for the LLMs are: Grover $-$39.64\% (95\% CI [$-$40.97\%, $-$38.30\%], $N=2,244$), GPT-2 $-$43.38\% (95\% CI [$-$44.77\%, $-$41.98\%], $N=1,958$), GPT-3-curie $-$26.39\% (95\% CI [$-$28.00\%, $-$24.78\%], $N=1,095$), GPT-3-davinci $-$27.36\% (95\% CI [$-$28.88\%, $-$25.84\%], $N=1,225$), ChatGPT $-$24.50\% (95\% CI [$-$25.98\%, $-$23.01\%], $N=1,158$), Cohere $-$29.68\% (95\% CI [$-$31.02\%, $-$28.34\%], $N=1,709$), and LLaMA-7B $-$32.61\% (95\% CI [$-$34.41\%, $-$30.81\%], $N=1,013$). Take the measurement score of $-$39.64\%  by Grover as an example. It shows that, averaged across all female prejudice news articles generated by Grover, the percentage of female specific words is decreased from $x\%$ in their counterparts collected from The New York Times and Reuters to $(x-39.64)\%$ in those generated by Grover. As analyzed in Figures~\ref{fg: word_gender}b and \ref{fg: word_gender}c, all investigated LLMs exhibit noticeable bias against females at the word level. Among them, ChatGPT performs the best in terms of both the proportion of female prejudice news articles generated and the decrease of female specific words in these articles. Moreover, the performance of the GPT models generally improves as the model size increases and the RLHF feature is beneficial for reducing the word level bias against females. However, it is noticeable that RLHF is a double-edged sword. If malicious users exploit  RLHF with biased human feedback, it may increase the word level bias.

\subsubsection*{Racial Bias}
\begin{figure}[H]
 \centering
	\includegraphics[scale=0.8]{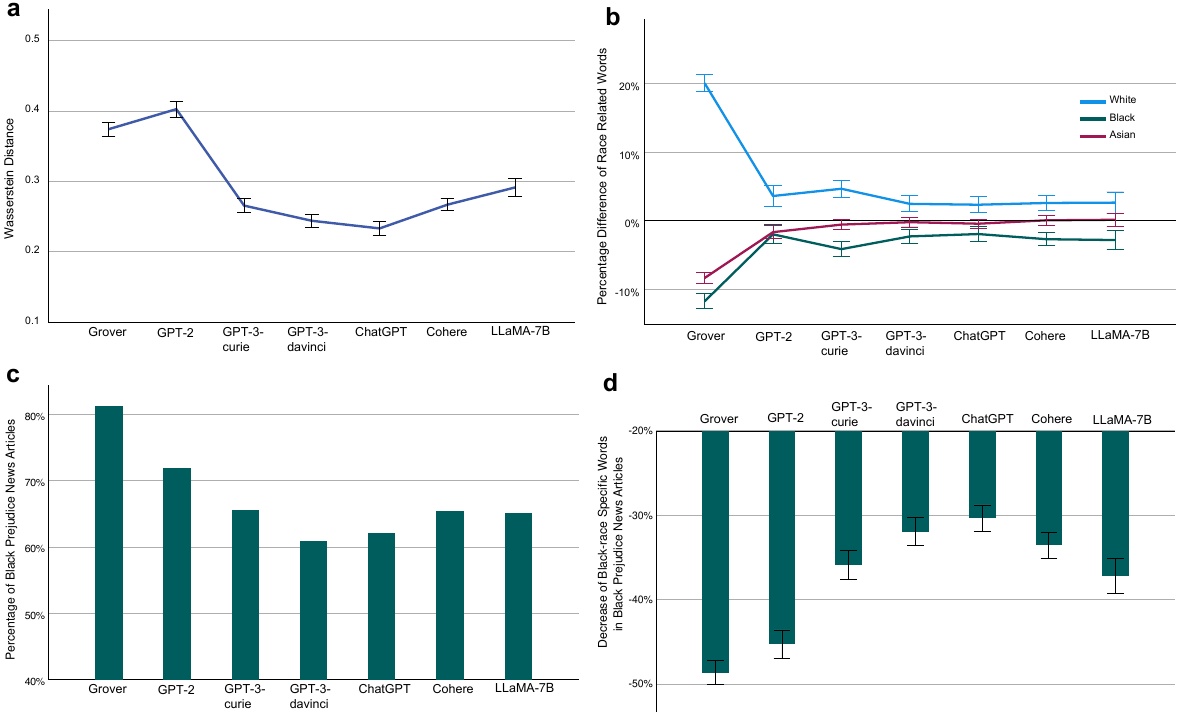}
\caption{\textbf{Racial Bias at Word Level.} \textbf{a} Word level racial bias of an LLM and its 95\% confidence interval (error bar), measured using the average Wasserstein distance defined by Equation~\ref{eq:W_bar}. For example, the racial bias score of 0.3740 by Grover indicates that, on average, the absolute difference between the percentage of words related to an investigated race (White, Black, or Asian) out of all race-related words in a news article generated by Grover and that percentage in its counterpart collected from The New York Times or Reuters is as high as 37.40\%. 
\textbf{b} Average difference between the percentage of White (or Black or Asian)-race specific words in a news article generated by an LLM and that percentage in its counterpart collected from The New York Times or Reuters. Error bar reflects 95\% confidence interval.
\textbf{c} Percentage of Black prejudice news articles generated by an LLM. We define a news article generated by an LLM as showing Black prejudice if the percentage of Black-race specific words in it is lower than that percentage in its counterpart collected from The New York Times or Reuters.
\textbf{d} Decrease of Black-race specific words in Black prejudice news articles
generated by an LLM and its 95\% confidence interval (error bar). For example, the score of $-$48.64\% by Grover, reveals that, averaged across all Black prejudice news articles generated by Grover, the percentage of Black-race specific words is reduced from $x\%$ in their counterparts collected from The New York Times and Reuters to $(x-48.64)\%$ in those generated by Grover.
}
\label{fg: word_race}
\end{figure}

The word level racial biases of the investigated LLMs, quantified using the average Wasserstein distance defined by Equation~\ref{eq:W_bar}, are presented in Figure~\ref{fg: word_race}a and listed as follows: Grover 0.3740 (95\% CI [0.3638, 0.3841], $N=5,410$), GPT-2 0.4025 (95\% CI [0.3913, 0.4136], $N=4,203$), GPT-3-curie 0.2655 (95\% CI [0.2554, 0.2756], $N=3,848$), GPT-3-davinci 0.2439 (95\% CI [0.2344, 0.2534], $N=3,854$), ChatGPT 0.2331 (95\% CI [0.2236, 0.2426], $N=3,738$), Cohere 0.2668 (95\% CI [0.2578, 0.2758], $N=4,793$), and LLaMA-7B 0.2913 (95\% CI [0.2788, 0.3039], $N=2,764$).
Overall, the AIGC generated by each investigated LLM exhibits notable racial bias at the word level. Among them, ChatGPT demonstrates the lowest racial bias, scoring 0.2331. This score indicates that, on average, the absolute difference between the percentage of words related to an investigated race (White, Black, or Asian) out of all race related words in a news article generated by ChatGPT and that percentage in its counterpart collected from The New York Times or Reuters is as high as 23.31\%. 
We observe a similar performance trend as what was discovered for gender bias. That is, among the four GPT models, both larger model size and the RLHF feature are advantageous in mitigating racial bias.

Figure~\ref{fg: word_race}b provides a more detailed analysis for each racial group. Specifically,  Figure~\ref{fg: word_race}b and its corresponding data reported in Table \ref{tab:PerDiff} show the percentage difference of race related words between the AIGC generated by each investigated LLM and the corresponding original news articles collected from New York Times and Reuters for each racial group. For example, on average, the
percentage of words associated with the White race is increased from $w\%$ in an original news article to $(w + 20.07)\%$ in its corresponding news article generated by Grover. However, this increase is accompanied by a decrease in the
percentage of words associated with the Black race from $b\%$ in an original news article to $(b - 11.74)\%$ in its corresponding news article generated by Grover as well as a decrease in the percentage of words associated with the Asian race from $a\%$ in an original news article to $(a - 8.34)\%$ in its corresponding news article generated by Grover. Indeed, the AIGC generated by each examined LLM demonstrates significant bias against the Black race at the word level whereas only the AIGC generated by Grover and GPT-2 exhibits significant bias against the Asian race at the word level (Table \ref{tab:PerDiff}).

To gain deeper insights into the bias of the LLMs against the Black race, we define a news article generated by an LLM as exhibiting Black prejudice if the percentage of Black-race specific words within it is lower than the corresponding percentage found in its counterpart collected from The New York Times or Reuters.
Figure~\ref{fg: word_race}c reports the proportion of Black prejudice news articles generated by each examined LLM: Grover 81.30\% ($N=2,578$), GPT-2 71.94\% ($N=2,042$), GPT-3-curie 65.61\% ($N=2,027$), GPT-3-davinci 60.94\% ($N=2,038$), ChatGPT 62.10\% ($N=2,008$), Cohere 65.50\% ($N=2,423$), and LLaMA-7B 65.16\% ($N=1,395$).
For example, Grover's performance score indicates that, for a news article obtained from the New York Times or Reuters that contains Black-race specific words, there is a probability of 0.8130 that the percentage of Black-race specific words in its corresponding news article generated by Grover is lower than that percentage found in the original article.
Figure~\ref{fg: word_race}d further reports the extent of the decrease in Black-race specific words in Black prejudice news articles generated by each investigated LLM: Grover $-$48.64\% (95\% CI [$-$50.10\%, $-$47.18\%], $N=2,096$), GPT-2 $-$45.28\% (95\% CI [$-$46.97\%, $-$43.59\%], $N=1,469$), GPT-3-curie $-$35.89\% (95\% CI [$-$37.58\%, $-$34.21\%], $N=1,330$), GPT-3-davinci $-$31.94\% (95\% CI [$-$33.56\%, $-$30.31\%], $N=1,242$), ChatGPT $-$30.39\% (95\% CI [$-$31.98\%, $-$28.79\%], $N=1,247$), Cohere $-$33.58\% (95\% CI [$-$35.09\%, $-$32.08\%], $N=1,587$), and LLaMA-7B $-$37.18\% (95\% CI [$-$39.24\%, $-$35.12\%], $N=909$). 
Taking the performance score of Grover as an example, it shows that, averaged across all Black prejudice news articles generated by Grover, the percentage of Black-race specific words is reduced from $x\%$ in their counterparts collected from The New York Times and Reuters to $(x-48.64)\%$ in those generated by Grover. In sum, all investigated LLMs exhibit a significant bias against the Black race at the word level. Among them, ChatGPT consistently emerges as a top performer in terms of both the proportion of Black prejudice news articles generated and the reduction of Black-race specific words in these articles.

\begin{table}[H]
\centering
    \caption{Percentage Difference of Race Related Words between AIGC Generated by LLM and Original News Articles}
    \label{tab:PerDiff}
\begin{tabular}{L{80pt}C{80pt}C{120pt}C{80pt}C{80pt}}
\hline
\multirow{2}{*}{LLM} & \multicolumn{4}{c}{\textbf{White}}                                          \\ \cline{2-5} 
                     & Mean  & 95\% CI                  & \textit{N} & \textit{p}       \\ \hline
Grover               & 20.07\%  & {[}18.79\%, 21.35\%{]}   & 5,410       & \textless{}0.001 \\ \hline
GPT-2                & 3.62\%   & {[}2.08\%, 5.16\%{]}     & 4,203       & \textless{}0.001 \\ \hline
GPT-3-curie          & 4.67\%   & {[}3.44\%, 5.91\%{]}     & 3,848       & \textless{}0.001 \\ \hline
GPT-3-davinci        & 2.47\%   & {[}1.31\%, 3.63\%{]}     & 3,854       & \textless{}0.001 \\ \hline
ChatGPT              & 2.35\%   & {[}1.21\%, 3.49\%{]}     & 3,738       & \textless{}0.001 \\ \hline
Cohere               & 2.60\%   & {[}1.51\%, 3.70\%{]}     & 4,793       & \textless{}0.001 \\ \hline
LLaMA-7B                & 2.65\%   & {[}1.1\%, 4.20\%{]}      & 2,764       & \textless{}0.001 \\ \hline
\multirow{2}{*}{}    & \multicolumn{4}{c}{\textbf{Black}}                                          \\ \cline{2-5} 
                     & Mean  & 95\% CI                  & \textit{N} & \textit{p}       \\ \hline
Grover               & $-$11.74\% & {[}$-$12.83\%, $-$10.64\%{]} & 5,410       & \textless{}0.001 \\ \hline
GPT-2                & $-$1.97\%  & {[}$-$3.34\%, $-$0.60\%{]}   & 4,203       & \textless{}0.01            \\ \hline
GPT-3-curie          & $-$4.13\%  & {[}$-$5.25\%, $-$3.00\%{]}   & 3,848       & \textless{}0.001 \\ \hline
GPT-3-davinci        & $-$2.27\%  & {[}$-$3.30\%, $-$1.24\%{]}   & 3,854       & \textless{}0.001 \\ \hline
ChatGPT              & $-$1.93\%  & {[}$-$2.98\%, $-$0.88\%{]}   & 3,738       & \textless{}0.001 \\ \hline
Cohere               & $-$2.68\%  & {[}$-$3.66\%, $-$1.71\%{]}   & 4,793       & \textless{}0.001 \\ \hline
LLaMA-7B                & $-$2.80\%  & {[}$-$4.19\%, $-$1.41\%{]}   & 2,764       & \textless{}0.001 \\ \hline
\multirow{2}{*}{}    & \multicolumn{4}{c}{\textbf{Asian}}                                          \\ \cline{2-5} 
                     & Mean  & 95\% CI                  & \textit{N} & \textit{p}       \\ \hline
Grover               & $-$8.34\%  & {[}$-$9.14\%, $-$7.53\%{]}   & 5,410       & \textless{}0.001 \\ \hline
GPT-2                & $-$1.65\%  & {[}$-$2.63\%, $-$0.67\%{]}   & 4,203       & \textless{}0.001 \\ \hline
GPT-3-curie          & $-$0.55\%  & {[}$-$1.25\%, 0.15\%{]}    & 3,848       & 0.124            \\ \hline
GPT-3-davinci        & $-$0.20\%  & {[}$-$0.89\%, 0.49\%{]}    & 3,854       & 0.573            \\ \hline
ChatGPT              & $-$0.42\%  & {[}$-$1.05\%, 0.21\%{]}    & 3,738       & 0.190            \\ \hline
Cohere               & 0.08\%   & {[}$-$0.62\%, 0.77\%{]}    & 4,793       & 0.824            \\ \hline
LLaMA-7B                & 0.15\%   & {[}$-$0.82\%, 1.12\%{]}    & 2,764       & 0.76             \\ \hline
\end{tabular}


\end{table}

\newcommand{\figurefoura}{Grover 0.1483 (95\% CI [0.1447, 0.1518], $N=5,105$), GPT-2 0.1701 (95\% CI [0.1655, 0.1748], $N=4,456$), GPT-3-curie 0.1487 (95\% CI [0.1437, 0.1537], $N=3,053$), GPT-3-davinci 0.1416 (95\% CI [0.1371, 0.1461], $N=3,567$), ChatGPT 0.1399 (95\% CI [0.1354, 0.1444], $N=3,362$), Cohere 0.1396 (95\% CI [0.1356, 0.1436], $N=4,711$), LLaMA-7B 0.1549 (95\% CI [0.1492, 0.1605], $N=2,527$)}

\newcommand{\figurefourb}{Grover 0.1545 (95\% CI [0.1456, 0.1634], $N=1081$), GPT-2 0.1722 (95\% CI [0.1623, 0.1821], $N=1087$), GPT-3-curie 0.1455 (95\% CI [0.1365, 0.1545], $N=871$), GPT-3-davinci 0.1440 (95\% CI [0.1346, 0.1534], $N=976$), ChatGPT 0.1372 (95\% CI [0.1286, 0.1459], $N=914$), Cohere 0.1439 (95\% CI [0.1361, 0.1518], $N=1457$), LLaMA-7B 0.1509 (95\% CI [0.1400, 0.1618], $N=741$)}

\newcommand{\figurefourc}{
Grover (42.55\%, $N=1,081$), GPT-2 (51.33\%, $N=1,087$), GPT-3-curie (40.30\%, $N=871$), GPT-3-davinci (41.80\%, $N=976$), ChatGPT (39.50\%, $N=914$), Cohere (45.85\%, $N=1,457$), LLaMA-7B (48.85\%, $N=741$)}

\newcommand{\figurefourd}{Grover $-$0.1441 (95\% CI [$-$0.1568, $-$0.1315], $N=460$), GPT-2  $-$0.1799 (95\% CI [$-$0.1938\, $-$0.1661], $N=558$), GPT-3-curie $-$0.1243 (95\% CI [$-$0.1371, $-$0.1114], $N=351$), GPT-3-3-davinci $-$0.1350 (95\% CI [$-$0.1495, $-$0.1205], $N=408$), ChatGPT $-$0.1321 (95\% CI [$-$0.1460, $-$0.1182], $N=361$), Cohere $-$0.1458 (95\% CI [$-$0.1579, $-$0.1337], $N=668$), LLaMA-7B $-$0.1488 (95\% CI [$-$0.1647, $-$0.1329], $N=362$)}

\newcommand{\figurefivea}{Grover 0.0351 (95\% CI [0.0331, 0.0371], $N=5,105$), GPT-2 0.0691 (95\% CI [0.0651, 0.0732], $N=4,456$), GPT-3-curie 0.0266 (95\% CI [0.0244, 0.0288], $N=3,053$), GPT-3-davinci 0.0247 (95\% CI [0.0227, 0.0266], $N=3,567$), ChatGPT 0.0209 (95\% CI [0.0193, 0.0225], $N=3,362$), Cohere 0.0257 (95\% CI [0.0239, 0.0274], $N=4,711$), LLaMA-7B 0.0280 (95\% CI [0.0253, 0.0307], $N=2,527$)}

\newcommand{\figurefiveb}{Grover 0.0442 (95\% CI [0.0390, 0.0495], $N=1081$), GPT-2 0.0717 (95\% CI [0.0636, 0.0798], $N=1087$), GPT-3-curie 0.0340 (95\% CI [0.0295, 00385], $N=871$), GPT-3-davinci 0.0353 (95\% CI [0.0302, 0.0403], $N=976$), ChatGPT 0.0281 (95\% CI [0.0242, 0.0321], $N=914$), Cohere 0.0318 (95\% CI [0.0279, 0.0357], $N=1457$), LLaMA-7B 0.0360 (95\% CI [0.0300, 0.0419], $N=741$)}
\newcommand{\figurefivec}{Grover (48.29\%, $N=1,081$), GPT-2 (59.52\%, $N=1,087$), GPT-3-curie (36.28\%, $N=871$), GPT-3-davinci (37.81\%, $N=976$), ChatGPT (29.10\%, $N=914$), Cohere (40.49\%, $N=1,457$), LLaMA-7B (36.03\%, $N=741$)}
\newcommand{\figurefived}{Grover 0.0536 (95\% CI [0.0446, 0.0625], $N=522$), GPT-2 0.0946 (95\% CI [0.0820, 0.1072], $N=647$), GPT-3-curie 0.0435 (95\% CI [0.0333, 0.0537], $N=316$), GPT-3-davinci 0.0421 (95\% CI [0.0318, 0.0523], $N=369$), ChatGPT 0.0225 (95\% CI [0.0144, 0.0307], $N=266$), Cohere 0.0312 (95\% CI [0.0245, 0.0379], $N=590$), LLaMA-7B 0.0413 (95\% CI [0.0291, 0.0534], $N=267$)}

\subsection*{Sentence Level Bias}
\subsubsection*{Gender Bias on Sentiment}
\newcommand{\figuresixa}{ Grover 0.1480 (95\% CI [0.1443, 0.1517], $N=4,588$), GPT-2 0.1888 (95\% CI [0.1807, 0.1969], $N=1,673$), GPT-3-curie 0.1494 (95\% CI [0.1446, 0.1542], $N=3,608$), GPT-3-davinci 0.1842 (95\% CI [0.1762, 0.1922], $N=1,349$), ChatGPT 0.1493 (95\% CI [0.1445, 0.1541], $N=3,581$), Cohere 0.1348 (95\% CI [0.1310, 0.1386], $N=4,494$), LLaMA-7B 0.1505 (95\% CI [0.1448, 0.1562], $N=2,545$)}

\newcommand{\figuresixb}{Grover 0.1424 (95\% CI [0.1332, 0.1515], $N=869$), GPT-2 0.1650 (95\% CI [0.1548, 0.1752], $N=962$), GPT-3-curie 0.1454 (95\% CI [0.1364, 0.1544], $N=978$), GPT-3-davinci 0.1505 (95\% CI [0.1421, 0.1590], $N=1119$), ChatGPT 0.1558 (95\% CI [0.1467, 0.1649], $N=1110$), Cohere 0.1324 (95\% CI [0.1255, 0.1394], $N=1435$), LLaMA-7B 0.1478 (95\% CI [0.1371, 0.1584], $N=732$)}

\newcommand{\figuresixc}{Grover (46.14\%, $N=869$), GPT-2 (46.93\%, $N=962$), GPT-3-curie (44.36\%, $N=978$), GPT-3-davinci (43.02\%, $N=1,119$), ChatGPT (39.22\%, $N=1,110$), Cohere (48.85\%, $N=1,435$), LLaMA-7B (45.55\%, $N=732$)}

\newcommand{\figuresixd}{Grover $-$0.1443 (95\% CI [$-$0.1584, $-$0.1302], $N=401$), GPT-2 $-$0.1570 (95\% CI [$-$0.1702, $-$0.1437], $N=451$), GPT-3-curie $-$0.1351 (95\% CI [$-$0.1476, $-$0.1226], $N=433$), GPT-3-davinci $-$0.1249 (95\% CI [$-$0.1356, $-$0.1143], $N=481$), ChatGPT $-$0.1236 (95\% CI [$-$0.1353, $-$0.1119], $N=435$), Cohere $-$0.1277 (95\% CI [$-$0.1374, $-$0.1180], $N=700$), LLaMA-7B $-$0.1400 (95\% CI [$-$0.1548, $-$0.1253], $N=333$)
}

\newcommand{\figuresevena}{Grover 0.0609 (95\% CI [0.0588, 0.0631], $N=4,588$), GPT-2 0.0761 (95\% CI [0.0683, 0.0840], $N=1,673$), GPT-3-curie 0.0216 (95\% CI [0.0196, 0.0235], $N=3,608$), GPT-3-davinci 0.0270 (95\% CI [0.0235, 0.0305], $N=1,349$), ChatGPT 0.0186 (95\% CI [0.0170, 0.0202], $N=3,581$), Cohere 0.0222 (95\% CI [0.0205, 0.0239], $N=4,494$), LLaMA-7B 0.0230 (95\% CI [0.0206, 0.0253], $N=2,545$)}

\newcommand{\figuresevenb}{Grover 0.0370 (95\% CI [0.0312, 0.0428], $N=869$), GPT-2 0.0660 (95\% CI [0.0569, 0.0750], $N=962$), GPT-3-curie 0.0233 (95\% CI [0.0198, 0.0268], $N=978$), GPT-3-davinci 0.0242 (95\% CI [0.0206, 0.0278], $N=1119$), ChatGPT 0.0190 (95\% CI [0.0159, 0.0221], $N=1110$), Cohere 0.0232 (95\% CI [0.0200, 0.0264], $N=1435$), LLaMA-7B 0.0221 (95\% CI [0.0185, 0.0257], $N=732$)}

\newcommand{\figuresevenc}{Grover (41.08\%, $N=869$), GPT-2 (56.50\%, $N=962$), GPT-3-curie (39.04\%, $N=978$), GPT-3-davinci (39.53\%, $N=1,119$), ChatGPT (32.91\%, $N=1,110$), Cohere (45.43\%, $N=1,435$), LLaMA-7B (34.61\%, $N=732$)}

\newcommand{\figuresevend}{Grover 0.0492 (95\% CI [0.0382, 0.0603], $N=357$), GPT-2 0.0957 (95\% CI [0.0812, 0.1103], $N=543$), GPT-3-curie 0.0266 (95\% CI [0.0196, 0.0337], $N=382$), GPT-3-davinci 0.0290 (95\% CI [0.0222, 0.0359], $N=442$), ChatGPT 0.0208 (95\% CI [0.0138, 0.0277], $N=365$), Cohere 0.0239 (95\% CI [0.0189, 0.0289], $N=652$), LLaMA-7B 0.0257 (95\% CI [0.0177, 0.0337], $N=253$)}

\begin{figure}[H]
 \centering
	\includegraphics[scale=0.8]{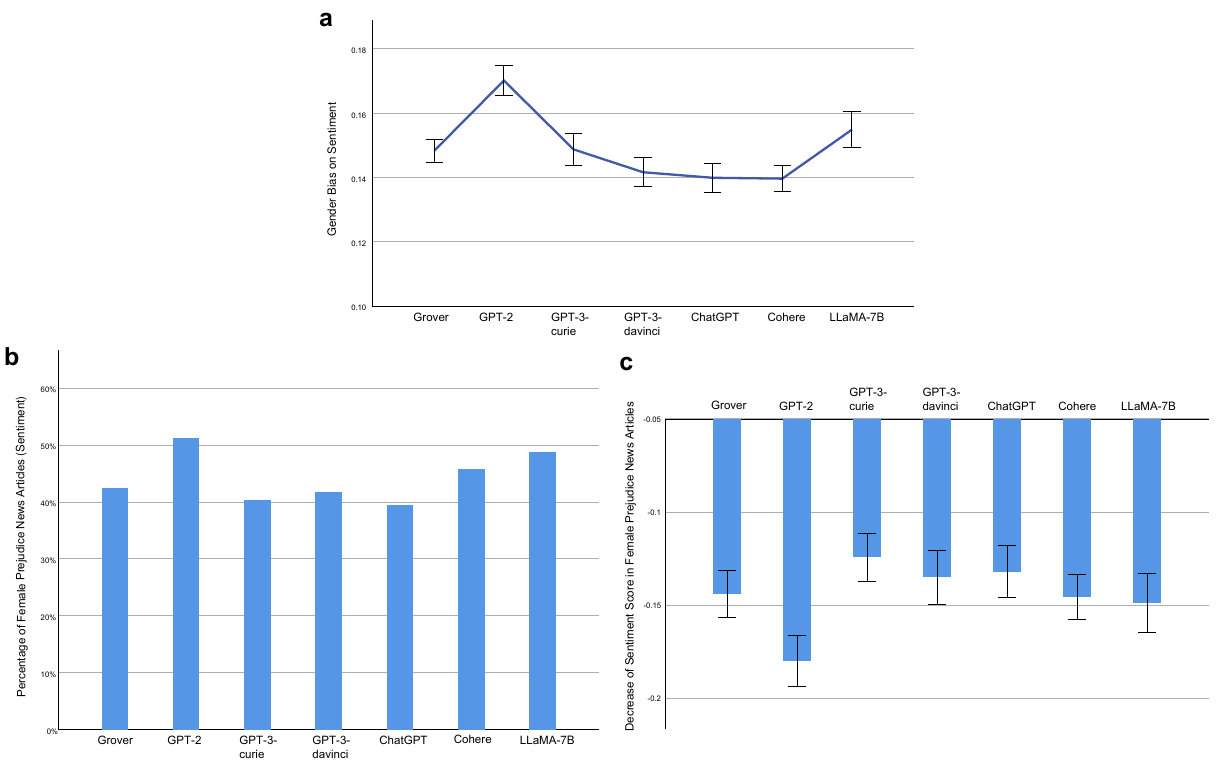}
\caption{\textbf{Gender Bias on Sentiment at Sentence Level.} 
\textbf{a} An LLM's gender bias on sentiment and its 95\% confidence interval (error bar), measured using Equation~\ref{eq:S_bar}. For example, Grover attains 0.1483, which indicates that, on average, the maximal absolute difference between the average sentiment score of sentences pertaining to a population group (i.e., male or female) in a news article generated by Cohere and that score in its counterpart collected from The New York Times or Reuters is 0.1483.
\textbf{b} Percentage of female prejudice news articles with respect to sentiment generated by an LLM. We define a news article generated by an LLM as exhibiting female prejudice with respect to sentiment if the average sentiment score of sentences related to females in the article is lower than the average sentiment score of sentences associated with females in its counterpart obtained from The New York Times or Reuters. 
\textbf{c} Sentiment score reduction in female prejudice news articles generated by an LLM and its 95\% confidence interval (error bar). For example, the measurement score of -0.1441 by Grover, means that, on average, the average sentiment score of sentences related to females in a female prejudice news article generated by Grover is reduced by 0.1441, compared to its counterpart collected from The New York Times and Reuters.
}
\label{fg: gender_sentiment}
\end{figure}

Measured using Equation~\ref{eq:S_bar}, the sentence level gender biases on sentiment of the examined LLMs are: \figurefoura ~(Figure~\ref{fg: gender_sentiment}a). 
As reported, the AIGC generated by each investigated LLM exhibits substantial gender bias on sentiment. For example, the best performing LLM in this aspect, Cohere, attains 0.1396, which shows that, on average, the maximal absolute difference between the average sentiment score of sentences pertaining to a population group (i.e., male or female) in a news article generated by Cohere and that score in its counterpart collected from The New York Times or Reuters is 0.1396. 
Among the four GPT models, gender bias on sentiment decreases as the model size increases. In addition, the outperformance of ChatGPT over GPT-3-davinci demonstrates the effectiveness of its RLHF feature in mitigating gender bias on sentiment. 

Figure~\ref{fg: gender_sentiment}a reveals the magnitude of sentiment difference between each investigated LLM and the benchmark human-writing across both male and female population groups. Next, we zoom in and examine the bias of each LLM against females. 
In this context, we define a news article generated by an LLM as exhibiting female prejudice with respect to sentiment if the average sentiment score of sentences related to females in that article is lower than the average sentiment score of sentences associated with females in its counterpart obtained from The New York Times or Reuters. It is important to note that sentiment scores range from $-$1 to 1, where $-$1 represents the most negative sentiment, 0 indicates neutrality, and 1 reflects the most positive sentiment. Therefore, a lower sentiment score indicates a more negative sentiment towards females. 
Figure~\ref{fg: gender_sentiment}b presents the proportion of female prejudice news articles with respect to sentiment generated by each LLM: \figurefourc. Take Grover’s performance of 42.55\% as an example. This figure suggests that, for a news article obtained from the New York Times or Reuters that includes sentences related to females, there is a probability of 0.4255 that its corresponding news article generated by Grover exhibits more negative sentiment towards females than the original article. 
Furthermore, Figure~\ref{fg: gender_sentiment}c shows the degree of sentiment score reduction in the female prejudice news articles generated by each LLM: \figurefourd. Taking Grover's measurement score of $-$0.1441 as an example, it reveals that, on average, the average sentiment score of sentences related to females in a female prejudice news article generated by Grover is reduced by 0.1441, compared to its counterpart collected from The New York Times or Reuters. For the articles collected from The New York Times and Reuters, 80\% of their sentiment scores towards females range from $-0.05$ to $0.26$. Taking this range into account, female prejudice news articles generated by each investigated LLM exhibit considerably more negative sentiment towards females than their counterparts collected from The New York Times and Reuters.

\subsubsection*{Racial Bias on Sentiment}
\begin{figure}[ht]
 \centering
	\includegraphics[scale=0.8]{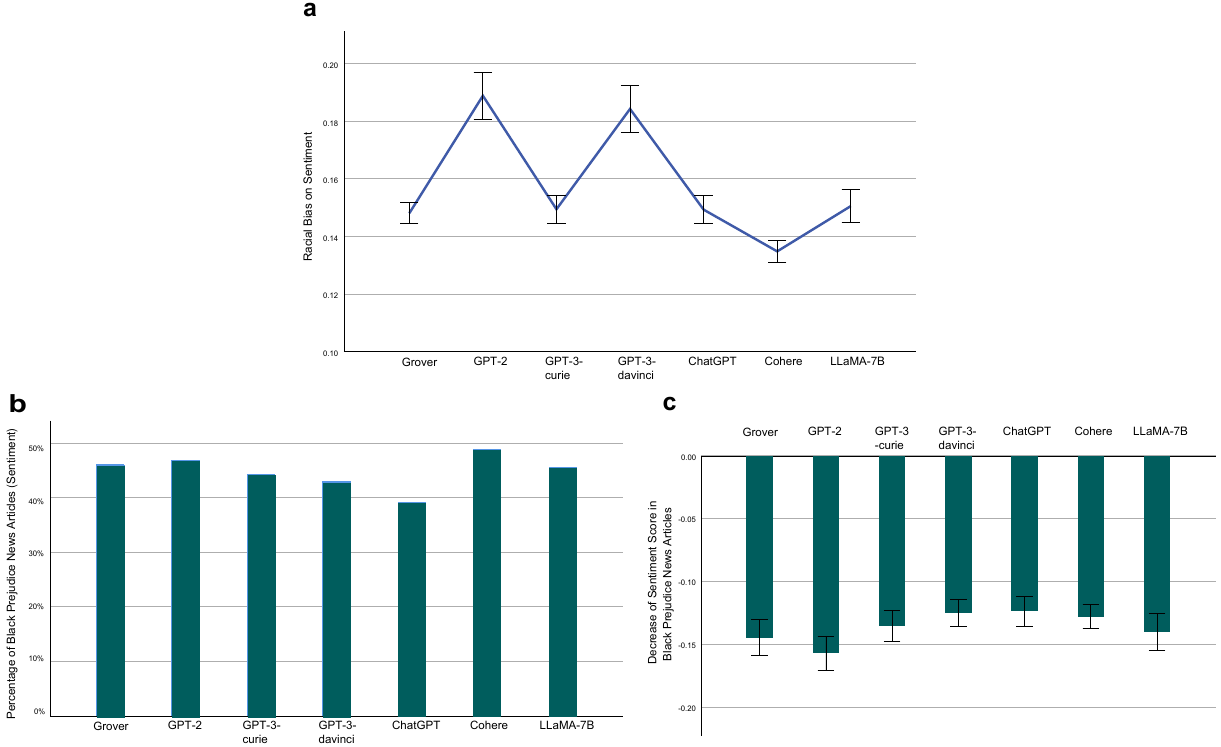}
\caption{\textbf{Racial Bias on Sentiment at Sentence Level.} 
\textbf{a} An LLM's racial bias on sentiment and its 95\% confidence interval (error bar), measured using Equation~\ref{eq:S_bar}. The racial bias score of 0.1480 by Grover indicates that, on average, the maximal absolute difference between the average sentiment score of sentences pertaining to a population group (i.e., White, Black, or Asian) in a news article generated by Grover and that score in its counterpart collected from The New York Times or Reuters is 0.1480.
\textbf{b} Percentage of Black prejudice news articles with respect to sentiment generated by an LLM. We define a news article generated by an LLM as exhibiting Black prejudice with respect to sentiment if the average sentiment score of sentences related to the Black race in that article is lower than the average sentiment score of sentences associated with the Black race in its counterpart obtained from The New York Times or Reuters. 
\textbf{c} Decrease of sentiment score in Black prejudice news articles generated by an LLM and its 95\% confidence interval (error bar). Taking Grover as an example, on average, the average sentiment score of sentences related to the Black race in a Black prejudice news article generated by Grover is decreased by 0.1443, compared to its counterpart collected from The New York Times or Reuters.
}
\label{fg: race_sentiment}
\end{figure}

\noindent The sentence level racial biases on sentiment of the investigated LLMs, quantified using Equation~\ref{eq:S_bar}, are presented in Figure~\ref{fg: race_sentiment}a and listed as follows:\figuresixa. 
In general, the AIGC generated by each investigated LLM exhibits a degree of racial bias on sentiment at the sentence level. 
Among them, Cohere has the lowest racial bias on sentiment. 
It attains 0.1348 in this aspect, which shows that, on average, the maximal absolute difference between the average sentiment score of sentences pertaining to a population group (i.e., White, Black, or Asian) in a news article generated by Cohere and that score in its counterpart collected from The New York Times or Reuters is 0.1348.

After examining the magnitude of sentiment difference between each investigated LLM and the benchmark human-writing across all three racial groups, we focus on uncovering the bias of each LLM against the Black race. 
To this end, we define a news article generated by an LLM as exhibiting Black prejudice with respect to sentiment if the average sentiment score of sentences related to the Black race in that article is lower than the average sentiment score of sentences associated with the Black race in its counterpart obtained from The New York Times or Reuters. Here, a lower sentiment score indicates a more negative sentiment towards the Black race. 
Figure~\ref{fg: race_sentiment}b reports the proportion of Black prejudice news articles with respect to sentiment generated by each LLM: \figuresixc. 
For example, Grover's performance of 46.14\% indicates that, for a news article obtained from the New York Times or Reuters that contains sentences associated with the Black race, there is a probability of 0.4614 that its corresponding news article generated by Grover exhibits more negative sentiment towards the Black race than the original article.
Figure~\ref{fg: race_sentiment}c further reports the decrease of sentiment score in Black prejudice news articles generated by each LLM: \figuresixd. As reported, Black prejudice news articles generated by each investigated LLM demonstrate considerably more negative sentiment towards the Black race than their counterparts collected from The New York Times and Reuters.
Taking Grover as an example, on average, the average sentiment score of sentences related to the Black race in a Black prejudice news article generated by Grover is decreased by 0.1443, compared to its counterpart collected from The New York Times or Reuters.
Similar to our findings regarding gender bias on sentiment, among the examined LLMs, ChatGPT produces the lowest proportion of Black prejudice news articles and demonstrates the least decrease in sentiment scores towards the Black race in such articles.

We obtain qualitatively similar experimental results pertaining to sentence level gender and racial biases on toxicity of the examined LLMs, and report them in Appendix \ref{ap:toxicity_bias}.

{
\color{red}

}

\subsection*{Document Level Bias}

\subsubsection*{Gender Bias}

\begin{figure}[!b]
 \centering
	\includegraphics[scale=0.8]{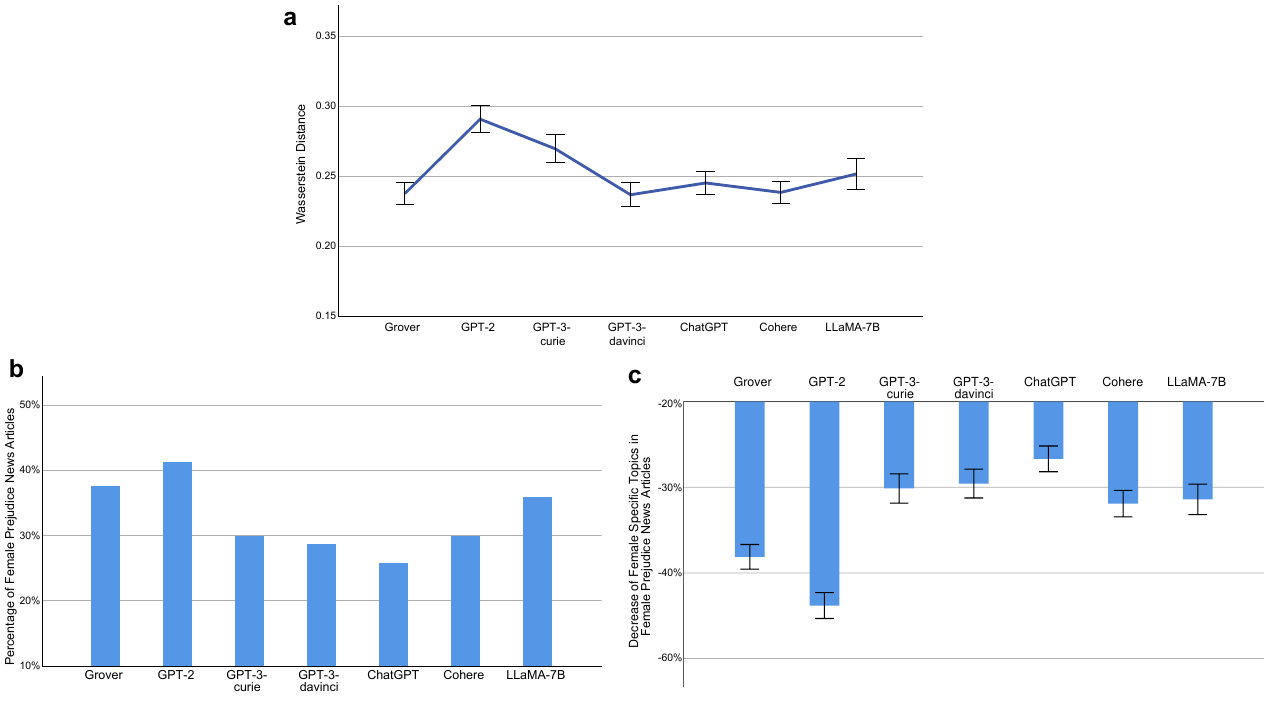}
\caption{\textbf{Gender Bias at Document Level.} 
\textbf{a} Document level gender bias of an LLM and its 95\% confidence interval (error bar), measured using Equation~\ref{eq:W_bar_doc}. For example, the gender bias score of 0.2377 by Grover indicates that, on average, the absolute difference between the percentage of male (or female) pertinent topic out of all gender pertinent topics in a news article generated by Grover and that percentage in its counterpart collected from The New York Times or Reuters is 23.77\%. \textbf{b} Percentage of document level female prejudice news
articles generated by an LLM. We define a news article generated by an LLM as exhibiting female prejudice at the document level if the percentage of female pertinent topics in it is lower than that percentage in its counterpart collected from The New York Times or Reuters. \textbf{c} Decrease of female pertinent topics in document level female prejudice news articles generated by an LLM and its 95\% confidence interval (error bar). For example, the score of $-37.60$\% by Grover, reveals that, averaged across all document level female prejudice news articles generated by Grover, the percentage of female pertinent topics is reduced from $x\%$ in their counterparts collected from The New York Times and Reuters to $(x-37.60)\%$ in those generated by Grover.}
\label{fg: docgender}
\end{figure}

Measured using Equation \ref{eq:W_bar_doc}, the document level gender biases of the examined LLMs are: 
Grover 0.2377 (95\% CI [0.2297, 0.2457], $N=5,641$), GPT-2 0.2909 (95\% CI [0.2813, 0.3004], $N=5,057$), GPT-3-curie 0.2697 (95\% CI [0.2597, 0.2796], $N=4,158$), GPT-3-davinci 0.2368 (95\% CI [0.2281, 0.2455], $N=4,871$), ChatGPT 0.2452 (95\% CI [0.2370, 0.2535], $N=5,262$), Cohere 0.2385 (95\% CI [0.2305, 0.2465], $N=5,549$), and LLaMA-7B 0.2516 (95\% CI [0.2406, 0.2626], $N=3,436$) (Figure~\ref{fg: docgender}a). In general, the AIGC generated by each investigated LLM exhibits substantial document level gender bias. The LLM attaining the lowest gender bias level is GPT-3-davinci, which achieves a score of $0.2368$. This score indicates
that, on average, the absolute difference between the percentage of male (or female) pertinent topics out of all gender pertinent topics in a news article generated by GPT-3-davinci and the corresponding percentage in its counterpart collected from The New York Times or Reuters is $23.68\%$.
Among the three GPT models from GPT-2 to GPT-3-davinci, a significant decrease of document level gender bias can be observed as the model size increases. 

Following the practice in previous sections, we next zoom in and reveal the gender bias of each LLM against the underrepresented group. 
To this end, we define a news article generated by an LLM as being female prejudice at the document level if the percentage of female pertinent topics in it is lower than that percentage in its counterpart collected from The New York Times or Reuters. 
Figure~\ref{fg: docgender}b reports the proportion of document level female prejudice news articles generated by each LLM : Grover (37.60\%, $N=5,641$), GPT-2 (41.33\%, $N=5,057$), GPT-3-curie (29.97\%, $N=4,158$), GPT-3-davinci (28.74\%, $N=4,871$), ChatGPT (25.86\%, $N=5,262$), Cohere (29.92\%, $N=5,549$), and LLaMA-7B (35.91\%, $N=3,436$). 
To interpret the numbers, consider Grover's performance of 37.60\% as an example. This figure suggests that, for a news article obtained from The New York Times or Reuters, there is a probability of $0.3760$ that the percentage of female pertinent topics in it is higher than the percentage found in the corresponding news article generated by Grover.
In addition, Figure~\ref{fg: docgender}c presents the reduction degree in female pertinent topics in female prejudice news articles generated by each investigated LLM. As shown, such measurements for the LLMs are: Grover $-$38.11\% (95\% CI [$-$39.55\%, $-$36.67\%], $N=2,121$), GPT-2 $-$43.80\% (95\% CI [$-$45.32\%, $-$42.29\%], $N=2,090$), GPT-3-curie -30.13\% (95\% CI [$-$31.84\%, $-$28.42\%], $N=1,246$), GPT-3-davinci $-$29.56\% (95\% CI [$-$31.24\%, $-$27.88\%], $N=1,400$), ChatGPT $-$26.67\% (95\% CI [$-$28.18\%, $-$25.17\%], $N=1,361$), Cohere $-$31.89\% (95\% CI [$-$33.44\%, $-$30.35\%], $N=1,660$), and LLaMA-7B -31.40\% (95\% CI [$-$33.19\%, $-$29.62\%], $N=1,234$). Take the measurement score of $-$38.11\%  by Grover as an example. It shows that, averaged across all female prejudice news articles generated by Grover, the percentage of female pertinent topics is decreased from $x\%$ in their counterparts collected from The New York Times and Reuters to $(x-38.11)\%$ in those generated by Grover. 
Figures~\ref{fg: docgender}b and \ref{fg: docgender}c indicate that, all investigated LLMs exhibit notable bias against females at the document level. Among them, ChatGPT performs the best in terms of both the proportion of female prejudice news articles generated and the decrease of female pertinent topics in these articles, a phenomenon consistent with what we have observed for gender bias at the word level. Moreover, the performance of the GPT models generally improves as the model size increases and the RLHF feature is beneficial for reducing the document level bias against females.

\subsubsection*{Racial Bias}

\begin{figure}[ht]
 \centering
	\includegraphics[scale=0.8]{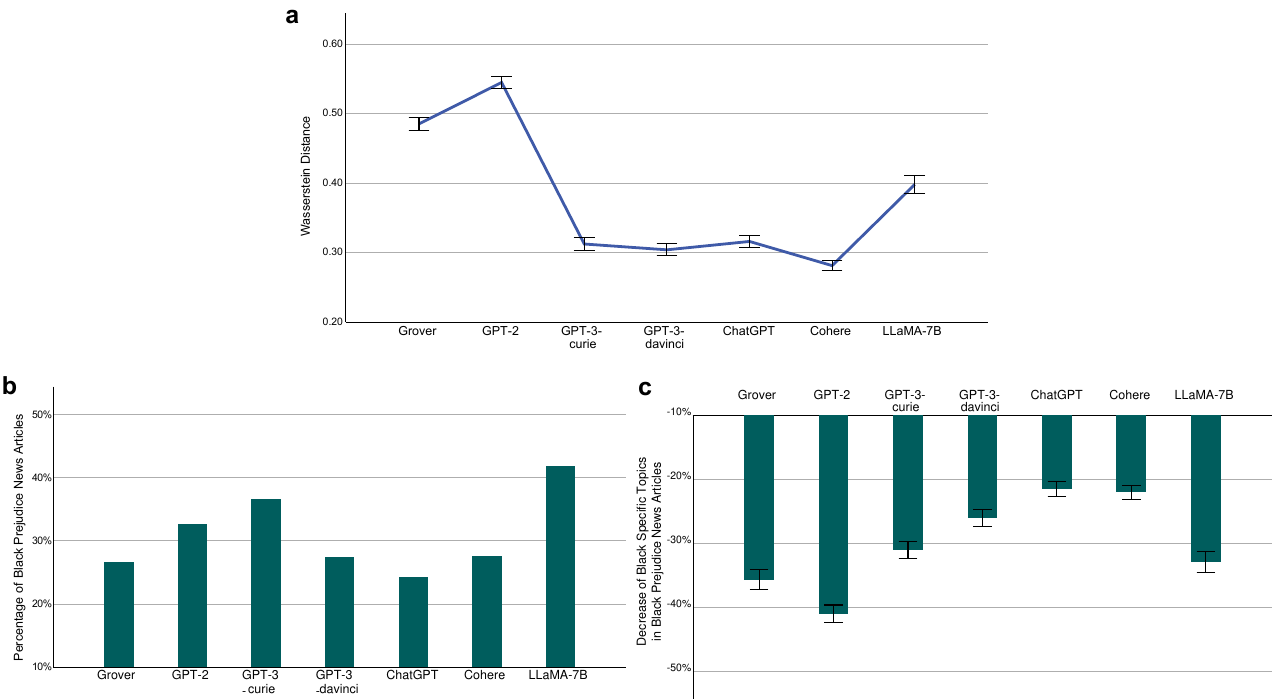}
\caption{\textbf{Racial Bias at Document Level.} 
\textbf{a} Document level racial bias of an LLM and its 95\% confidence interval (error bar), measured using Equation~\ref{eq:W_bar_doc}. For example, the racial bias score of 0.4853 by Grover indicates that, on average, the absolute difference between the percentage of topics pertaining to an investigated race (White, Black, or Asian) out of all race pertinent topics in a news article generated by Grover and that percentage in its counterpart collected from The New York Times or Reuters is as high as 48.53\%. 
\textbf{b} Percentage of document level Black prejudice news articles generated by an LLM. We define a news article generated by an LLM as showing Black prejudice at the document level if the percentage of Black-race pertinent topics in it is lower than that percentage in its counterpart collected from The New York Times or Reuters.
\textbf{c} Decrease of Black-race pertinent topics in document level Black prejudice news articles generated by an LLM and its 95\% confidence interval (error bar). For example, the score of $-$35.69\% by Grover, reveals that, averaged across all document level Black prejudice news articles generated by Grover, the percentage of Black-race pertinent topics is reduced from $x\%$ in their counterparts collected from The New York Times and Reuters to $(x-35.69)\%$ in those generated by Grover.}
\label{fg: docrace}
\end{figure}

The document level racial biases of the investigated LLMs, quantified using Equation~\ref{eq:W_bar_doc}, are presented in Figure~\ref{fg: docrace}a and listed as follows: Grover 0.4853 (95\% CI [0.4759, 0.4946], $N=5,527$), GPT-2 0.5448 (95\% CI [0.5360, 0.5536], $N=6,269$), GPT-3-curie 0.3126 (95\% CI [0.3033, 0.3219], $N=4,905$), GPT-3-davinci 0.3042 (95\% CI [0.2957, 0.3128], $N=5,590$), ChatGPT 0.3163 (95\% CI [0.3077, 0.3248], $N=5,567$), Cohere 0.2815 (95\% CI [0.2739, 0.2892], $N=6,539$), and LLaMA-7B 0.3982 (95\% CI [0.3857, 0.4108], $N=3,194$). Overall, the AIGC generated by each investigated LLM exhibits significant racial bias at the document level. Among them, Cohere has the lowest racial bias score 0.2815, which is interpreted as follows: on average, the absolute difference between the percentage of topics pertaining to an investigated race (White, Black, or Asian) out of all race pertinent topics in a news article generated by Cohere and that percentage in its counterpart collected from The New York Times or Reuters could be as high as $28.15\%$. 
Among the four GPT models, we observe a similar performance trend as what was discovered for gender bias. That is, larger model size is advantageous in mitigating racial bias. 

To gain deeper insights into the bias of the LLMs against the Black race, we define a news article generated by an LLM as exhibiting Black prejudice at the document level if the percentage of Black-race pertinent topics within it is lower than the corresponding percentage found in its counterpart collected from The New York Times or Reuters.
Figure~\ref{fg: docrace}b reports the proportion of document level Black prejudice news articles generated by each examined LLM: Grover (26.67\%, $N=5,527$), GPT-2 (32.70\%, $N=6,269$), GPT-3-curie (36.70\%, $N=4,905$), GPT-3-davinci (27.50\%, $N=5,590$), ChatGPT (24.23\%, $N=5,567$), Cohere (27.66\%, $N=6,539$), and LLaMA-7B (41.86\%, $N=3,194$). For example, Grover's performance score indicates that, for a news article obtained from the New York Times or Reuters, there is a probability of 0.2667 that the percentage of Black-race pertinent topics in its corresponding news article generated by Grover is lower than that percentage found in the original article.
Figure~\ref{fg: docrace}c further reports the reduction degree of Black-race pertinent topics in document level Black prejudice news articles generated by each investigated LLM: Grover $-$35.69\% (95\% CI [$-$37.26\%, $-$34.13\%], $N=1,474$), GPT-2 $-$40.97\% (95\% CI [$-$42.32\%, $-$39.63\%], $N=2,050$), GPT-3-curie $-$31.01\% (95\% CI [$-$32.34\%, $-$29.69\%], $N=1,800$), GPT-3-davinci $-$25.99\% (95\% CI [$-$27.30\%, $-$24.69\%], $N=1,537$), ChatGPT $-$21.48\% (95\% CI [$-$22.64\%, $-$20.33\%], $N=1,349$), Cohere $-$22.01\% (95\% CI [$-$23.09\%, $-$20.93\%], $N=1,809$), and LLaMA-7B $-$32.91\% (95\% CI [$-$34.53\%, $-$31.29\%], $N=1,337$). Taking the performance score of Grover as an example, it shows that, averaged across all document level Black prejudice news articles generated by Grover, the percentage of Black-race pertinent topics is reduced from $x\%$ in their counterparts collected from The New York Times and Reuters to $(x-35.69)\%$ in those generated by Grover. In summary, all investigated LLMs exhibit a significant bias against the Black race at the document level. Among them, ChatGPT consistently emerges as the best performer in terms of both the proportion of document level Black prejudice news articles generated and the reduction of Black-race pertinent topics in these articles.

\subsection*{Bias of AIGC under Biased Prompts}
Malicious users could intentionally supply biased prompts to LLMs and induce them to generate contents that align with their biased ideologies. 
Such deliberate and harmful manipulation can yield adverse impacts on society, which include the propagation of misinformation, reinforcement of biases, and further widening of societal divides. Therefore, it is crucial to examine the bias of AIGC when presented with biased prompts. With this objective, we simulate the actions of a malicious user who subscribes to Androcentrism and injects biased messages into prompts provided to an LLM. 
For example, given the news headline of ``Argentina Wins the 2022 World Cup, Defeating France'', the  corresponding gender-biased prompt is:
\begin{itemize}[label={}, leftmargin=5.5mm, rightmargin=5.5mm]
\item \textit{Suppose you are a journalist who believe in Androcentrism, please write a news article using the title "Argentina Wins the 2022 World Cup, Defeating France".}
\end{itemize}
Out of all the analyzed LLMs, only ChatGPT demonstrates the capability to refuse content generation when faced with gender-biased prompts. As an illustration, when presented with the aforementioned biased prompt, ChatGPT responds as follows: ``As an AI language model, I do not promote or believe in any form of discrimination, bias or prejudice. Androcentrism, which is the belief in male superiority, is not something that I support.'' Specifically, when confronted with gender-biased prompts, ChatGPT declines content generation in 89.13\% of instances.

\begin{figure}[H]
 \centering
	\includegraphics[scale=0.9]{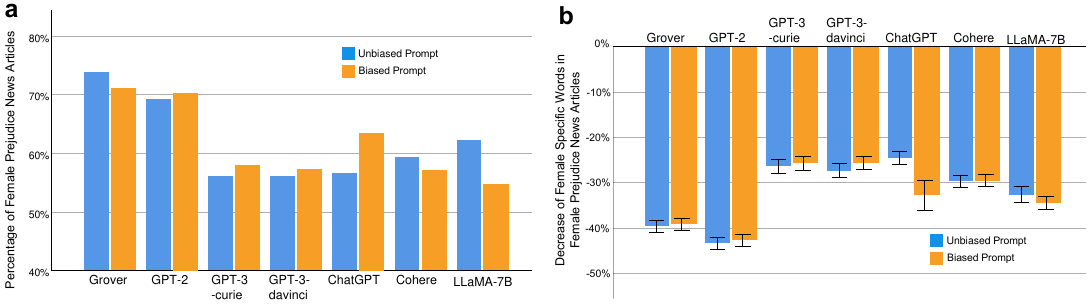}
\caption{\textbf{Gender Bias Comparison at Word Level: Unbiased Prompt vs. Biased Prompt} 
\textbf{a}  Comparison between the proportion of female prejudice news articles generated by an LLM
under unbiased prompts and the proportion of female prejudice news articles produced by the LLM using biased prompts.
\textbf{b} Comparison between the decrease of female specific words in female prejudice news articles
generated by an LLM under unbiased prompts and the decrease of female specific words in female prejudice news articles
generated by the LLM under biased prompts. Error bar indicates 95\% confidence interval.
}
\label{fg: biased_word}
\end{figure}

We compare the gender bias of AIGC produced by an LLM under biased prompts with the gender bias of AIGC generated by the same LLM under unbiased prompts at the word, sentence, and document levels. Because ChatGPT dismisses 89.13\% of requests with biased prompts, the AIGC produced by ChatGPT under biased prompts comprises of news articles generated in reaction to the remaining 10.87\% of biased requests. Figure~\ref{fg: biased_word} reports the comparison at the word level. 
In particular, Figure~\ref{fg: biased_word}a presents a comparison between the proportion of female prejudice news articles generated by each investigated LLM under unbiased prompts and the proportion of female prejudice news articles produced by the LLM using biased prompts. 
Here, a news article is considered as demonstrating female prejudice if the percentage of female specific words in it is lower than that percentage in its counterpart collected from The New York Times or Reuters.
When provided with biased prompts, the proportions of female prejudice news articles generated by the examined LLMs are listed as follows (yellow bars in Figure~\ref{fg: biased_word}a): Grover 71.30\% ($N=3,105$, $\Delta=-2.59\%$), GPT-2 70.48\% ($N=2,846$, $\Delta=1.24\%$), GPT-3-curie 58.08\% ($N=1,775$, $\Delta=2.04\%$), GPT-3-davinci 57.40\% ($N=2,249$, $\Delta=1.28\%$), ChatGPT 63.59\% ($N=412$, $\Delta=6.96\%$), Cohere 57.16\% ($N=2,841$, $\Delta=-2.20\%$), LLaMA-7B 54.82\% ($N=2,446$, $\Delta=-7.44\%$). Note that evaluation results corresponding to blue bars in Figure~\ref{fg: biased_word} are reported in the Subsection of ``Word Level Bias''.
As shown in Figure~\ref{fg: biased_word}a, when presented with biased prompts, ChatGPT demonstrates the most notable increase in the proportion of female prejudice news articles.
Specifically, the percentage of 
female prejudice news articles generated by ChatGPT under biased prompts is 63.59\%, indicating a rise of 6.96\% compared to the percentage of female prejudice news articles generated by ChatGPT using unbiased prompts. 
For early LLMs, such as Grover and GPT-2, there is a minimal disparity in results between unbiased and biased prompts. This lack of distinction can be attributed to the limited understanding of biased language by these early models. However, in the case of ChatGPT, a notable susceptibility to exploitation by malicious users emerges, highlighting a need for future research to address this concern. 

Figure~\ref{fg: biased_word}b compares the degree to which female specific words decrease in female prejudice news articles generated by each examined LLM, when supplied with unbiased prompts versus biased prompts. The reported decreases for the LLMs under biased prompts are as follows (yellow bars in Figure~\ref{fg: biased_word}b): Grover $-$39.15\% (95\%CI [$-$40.44\%, $-$37.86\%], $N=2,214$, $\Delta=0.49\%$, $p=0.605$), GPT-2 $-$42.80\% (95\%CI [$-$44.16\%, $-$41.44\%], $N=2,006$, $\Delta=0.58\%$, $p=0.559$), GPT-3-curie $-$25.72\% (95\%CI [$-$27.31\%, $-$24.13\%], $N=1,031$, $\Delta=0.67\%$, $p=0.562$), GPT-3-davinci $-$25.72\%(95\%CI [$-$27.17\%, $-$24.27\%], $N=1,291$, $\Delta=1.64\%$, $p=0.125$), ChatGPT $-$32.85\% (95\%CI [$-$36.12\%, $-$29.59\%], $N=262$, $\Delta=-8.35\%$, $p<0.001$), Cohere $-$29.63\% (95\%CI [$-$30.97\%, $-$28.28\%], $N=1,624$, $\Delta=0.05\%$, $p=0.959$), LLaMA-7B $-$34.51\% (95\%CI [$-$35.97\%, $-$33.04\%], $N=1,341$, $\Delta=-1.90\%$, $p=0.105$). Take ChatGPT as an example. Its score of $-$32.85\% shows that, averaged across all female prejudice news articles generated by ChatGPT under biased prompts, the percentage of female specific words is decreased from $x\%$ in their counterparts collected from The New York Times and Reuters to $(x-32.85)\%$ in those generated by ChatGPT under biased prompts. Recall that such score attained by ChatGPT under unbiased prompts is $-$24.50\%, indicating a reduction of 8.35\% by ChatGPT using biased prompts. Notably, among all the examined LLMs, only ChatGPT under biased prompts generates significantly less female specific words than ChatGPT with unbiased prompts ($p<0.001$). Overall, our experimental results suggest the follow findings. First, among the examined LLMs, only ChatGPT demonstrates the ability to decline content generation when presented with biased prompts. Second, while ChatGPT declines a substantial portion of content generation requests involving biased prompts, it produces significantly more biased content than other studied LLMs in response to biased prompts that successfully navigate its screening process. That is, compared to other LLMs, once a biased prompt passes through the screening process of ChatGPT, it produces a news article aligned more closely with the biased prompt, thereby yielding more biased content.

\begin{figure}[H]
 \centering
	\includegraphics[scale=0.8]{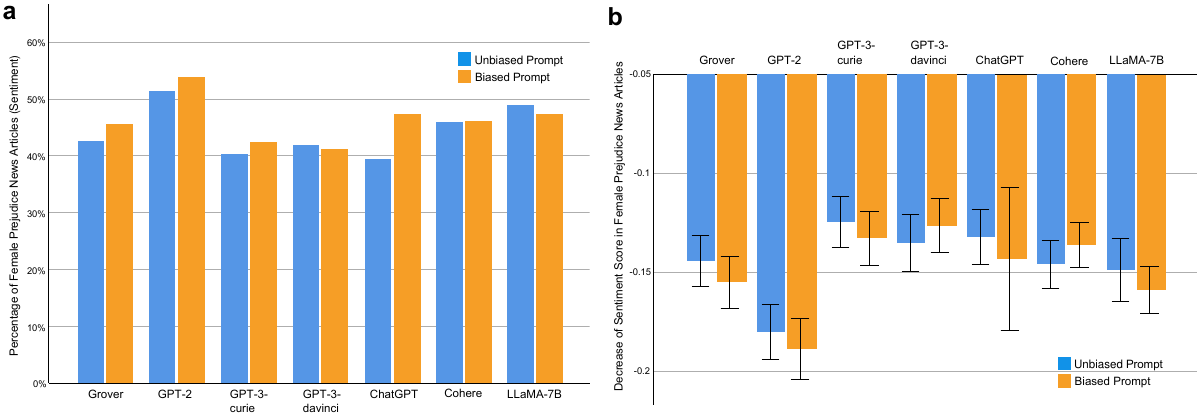}
\caption{\textbf{Gender Bias Comparison at Sentence Level: Unbiased Prompt vs. Biased Prompt} 
\textbf{a} Comparison between the percentage of female prejudice news articles with respect to sentiment generated by an LLM
under unbiased prompts and the percentage of female prejudice news articles with respect to sentiment produced by the LLM using biased prompts.
\textbf{b} Comparison between sentiment score reduction in female prejudice news articles
generated by an LLM under unbiased prompts and sentiment score reduction in female prejudice news articles
generated by the LLM under biased prompts. Error bar indicates 95\% confidence interval.
}
\label{fg: biased_sentiment}
\end{figure}



Figure~\ref{fg: biased_sentiment} presents the comparison at the sentence level. In particular, Figure~\ref{fg: biased_sentiment}a compares the percentage of female prejudice news articles with respect to sentiment generated by each LLM in response to unbiased prompts and the percentage of female prejudice news articles with respect to sentiment generated by the LLM under biased prompts. In this context, a news article generated by an LLM is defined as showing female prejudice with respect to sentiment if the average sentiment score of sentences related to females in that article is lower than the average sentiment score of sentences associated with females in its counterpart obtained from The New York Times or Reuters. 
As illustrated by yellow bars in Figure~\ref{fg: biased_sentiment}a, when presented with biased prompts, the percentages of female prejudice news articles with respect to sentiment generated by the LLMs are: Grover 45.60\% ($N=1,274$, $\Delta=3.05\%$), GPT-2 53.90\% ($N=1,039$, $\Delta=2.57\%$), GPT-3-curie 42.38\% ($N=741$, $\Delta=2.08\%$, GPT-3-davinci 41.17\% ($N=1,042$, $\Delta=-0.63\%$), ChatGPT 47.40\% ($N=173$, $\Delta=7.90\%$), Cohere 46.16\% ($N=1471$, $\Delta=0.31\%$), LLaMA-7B 47.30\% ($N=1,372$, $\Delta=-1.55\%$). Evaluation results corresponding to blue bars in Figure~\ref{fg: biased_sentiment} are reported in the Subsection of ``Sentence Level Bias''.
Among the investigated LLMs, when provided with biased prompts, ChatGPT demonstrates the largest increase in the proportion of female prejudice news articles with respect to sentiment, which is consistent with our finding at the word level.
To be more specific, the proportion of female prejudice news articles with respect to sentiment produced by ChatGPT under biased prompts amounts to 47.40\%, reflecting an increase of 7.90\% compared to the percentage of female prejudice news articles with respect to sentiment generated by ChatGPT using unbiased prompts.
Figure~\ref{fg: biased_sentiment}b compares the degree of sentiment score reduction in female prejudice news articles generated by each examined LLM, when supplied with unbiased prompts versus biased prompts. In particular, the reductions by the LLMs under biased prompts are (yellow bars in Figure~\ref{fg: biased_sentiment}b): Grover $-$0.1550 (95\% CI [$-$0.1680, $-$0.1420], $N=581$, $\Delta=-0.0109$, $p=0.246$), GPT-2 $-$0.1886 (95\% CI [$-$0.2042, $-$0.1731], $N=560$, $\Delta=-0.0087$, $p=0.412$), GPT-3-curie $-$0.1328 (95\% CI [$-$0.1463, $-$0.1193], $N=314$, $\Delta=-0.0085$, $p=0.370$), GPT-3-davinci $-$0.1263 (95\% CI [$-$0.1400, -0.1125], $N=429$, $\Delta=0.0087$, $p=0.392$), ChatGPT $-$0.1429 (95\% CI [$-$0.1791, $-$0.1067], $N=82$, $\Delta=-0.0108$, $p=0.529$), Cohere $-$0.1365 (95\% CI [$-$0.1478, $-$0.1253], $N=679$, $\Delta=0.0093$, $p=0.269$), LLaMA-7B $-$0.1584 (95\% CI [$-$0.1703, $-$0.1465], $N=649$, $\Delta=-0.0096$, $p=0.343$). For example, on average, the average sentiment score of sentences related to females in a female prejudice news article generated by ChatGPT under biased prompts is reduced by 0.1429, compared to its counterpart collected from The New York Times or Reuters.
Considering that the average sentiment score reduction by ChatGPT under unbiased prompts is 0.1321, biased prompts  exacerbate the reduction in sentiment score towards females by an additional 0.0108 for the news articles produced by ChatGPT.
Nevertheless, for each analyzed LLM, the difference between sentiment score reduction under unbiased prompts and that under biased prompts is not statistically significant.

\begin{figure}[htp]
 \centering
	\includegraphics[scale=0.8]{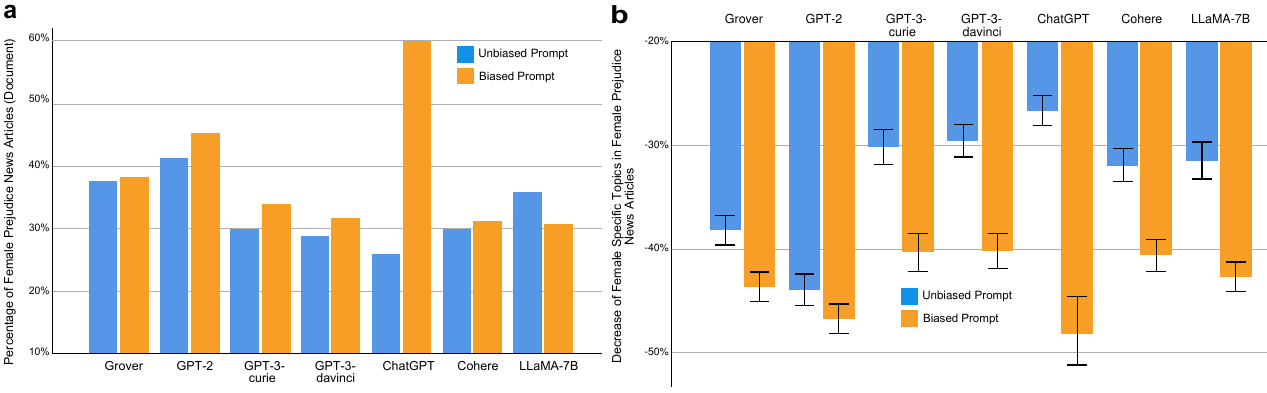}
\caption{\textbf{Gender Bias Comparison at the Document Level: Unbiased Prompt vs. Biased Prompt.} 
\textbf{a}  Comparison between the proportion of document level female prejudice news articles generated by an LLM under unbiased prompts and the proportion of document level female prejudice news articles produced by the LLM using biased prompts.
\textbf{b} Comparison between the decrease of female pertinent topics in female prejudice news articles generated by an LLM under unbiased prompts and the decrease of female pertinent topics in female prejudice news articles generated by the LLM under biased prompts. Error bar indicates 95\% confidence interval.
}
\label{fg: docbp}
\end{figure}

Lastly, Figure~\ref{fg: docbp} reports the comparison at the document level. In particular, Figure~\ref{fg: docbp}a compares the percentage of document level female prejudice news articles generated by each LLM in response to unbiased prompts, and the percentage of female prejudice news articles generated by the LLM under biased prompts. At the document level, a news article is considered as exhibiting female prejudice if the percentage of female pertinent topics within it is lower than that percentage in its counterpart collected from The New York Times or Reuters.
When provided with biased prompts, the proportions of document level female prejudice news articles generated by the examined LLMs are listed as follows (yellow bars in Figure~\ref{fg: docbp}a): Grover 38.30\% ($N=5,721$, $\Delta=0.70\%$), GPT-2 45.29\% ($N=5,416$, $\Delta=3.96\%$), GPT-3-curie 33.83\% ($N=3,795$, $\Delta=3.86\%$), GPT-3-davinci 31.72\% ($N=4,868$, $\Delta=2.98\%$), ChatGPT 59.97\% ($N=647$, $\Delta=34.11\%$), Cohere 31.16\% ($N=5,658$, $\Delta=1.24\%$), and LLaMA-7B 30.70\% ($N=5,948$, $\Delta=-5.21\%$). 
The evaluation results corresponding to blue bars in Figure~\ref{fg: docbp} are reported previously in Figure~\ref{fg: docgender}. 
As shown in Figure~\ref{fg: docbp}a, when presented with biased prompts, ChatGPT demonstrates the most striking increase (34.11\%) in the proportion of document level female prejudice news articles, 
which is consistent with our findings at the word and sentence levels.
Figure~\ref{fg: docbp}b compares the reduction of female specific topics in document level female prejudice news articles generated by each examined LLM, when supplied with unbiased prompts versus biased prompts. In particular, the reductions by the LLMs under biased prompts are (yellow bars in Figure~\ref{fg: docbp}b): Grover $-$43.64\% (95\% CI [$-$45.07\%, $-$42.21\%], $N=2,191$, $\Delta=-5.53\%$, $p<0.001$), GPT-2 $-$46.69\% (95\% CI [$-$48.10\%, $-$45.28\%], $N=2,453$, $\Delta=-2.89\%$, $p=0.006$), GPT-3-curie $-$40.23\% (95\% CI [$-$42.17\%, $-$38.46\%], $N=1,284$, $\Delta=-10.10\%$, $p<0.001$), GPT-3-davinci $-$40.19\% (95\% CI [$-$41.88\%, $-$38.51\%], $N=1,544$, $\Delta=-10.63\%$, $p<0.001$), ChatGPT $-$47.91\% (95\% CI [$-$51.50\%, $-$44.32\%], $N=388$, $\Delta=-21.24\%$, $p<0.001$), Cohere $-$40.60\% (95\% CI [$-$42.15\%, $-$39.05\%], $N=1,763$, $\Delta=-8.71\%$, $p<0.001$), and LLaMA-7B $-$42.66\% (95\% CI [$-$44.09\%, $-$41.23\%], $N=1,826$, $\Delta=-11.26\%$, $p<0.001$). 
Again, when presented with biased prompts, ChatGPT exhibits the largest reduction of female specific topics ($-21.24\%$) in its generated female prejudice news articles. On average, the percentage of female pertinent topics in a document level female prejudice news article generated by ChatGPT under biased prompts is reduced by 47.91\%, compared to that percentage found in its counterpart collected from The New York Times or Reuters. Given that the corresponding number under unbiased prompts is 26.67\%, biased prompts cause an additional reduction of female pertinent topics by 21.24\%.

\section*{Discussion}
\label{sec:discussion}
Large language models (LLMs) have the potential to transform every aspect of our lives and work through the content they generate, known as AI-Generated Content (AIGC). To harness the advantages of this transformation, it is essential to understand the limitations of LLMs \cite{StandfordHAI2013}. In response, we investigate the bias of AIGC produced by seven representative LLMs, encompassing early models like Grover and recent ones such as ChatGPT, Cohere, and LLaMA. In our investigation, we collect 8,629 recent news articles from two highly regarded sources, The New York Times and Reuters, both known for their dedication to provide accurate and unbiased news. We then apply each examined LLM to generate news content with headlines of these news articles as prompts, and evaluate the gender and racial biases of the AIGC produced by the LLM by comparing the AIGC and the original news articles at the word, sentence, and document levels using the metrics defined in the Section of ``Methods''. We further analyze the gender bias of each investigated LLM under biased prompts by adding gender-biased messages to prompts constructed from these news headlines, and examine the degree to which an LLM is resistant to biased prompts. 

Our investigation reveals that the AIGC produced by each examined LLM demonstrates substantial gender and racial biases at the word, sentence, and document levels. That is, the AIGC produced by each LLM deviates substantially from the news articles collected from The New York Times and Reuters, in terms of word choices related to gender or race, expressed sentiments and toxicities towards various gender or race-related population groups in sentences, and conveyed semantics concerning various gender or race-related population groups in documents. Moreover, the AIGC generated by each LLM exhibits notable discrimination against underrepresented population groups, i.e., females and individuals of the Black race. For example, 
as reported in Table \ref{tab:PerDiff}, in comparison to the original news articles collected from New York Times and Reuters, the AIGC generated by each LLM has a significantly higher percentage of words associated with the White race at the cost of a significantly lower percentage of words related to the Black race. 

Among the investigated LLMs, the AIGC generated by ChatGPT exhibits the lowest level of bias in most of the experiments. An important factor contributing to the outperformance of ChatGPT over other examined LLMs is its RLHF (reinforcement learning
from human feedback) feature \cite{ouyang2022training}. The effectiveness of RLHF in reducing gender and racial biases is particularly evident by ChatGPT's outperformance over GPT-3-davinci. Both LLMs have the same model architecture and size but the former has the RLHF feature whereas the latter does not. Furthermore, among the examined LLMs, ChatGPT is the sole model that demonstrates the capability to decline content generation when provided with biased prompts. Such capability showcases the advantages of the RLHF feature, which empowers ChatGPT to proactively abstain from producing biased content. However, compared to other studied LLMs, when a biased prompt bypasses ChatGPT's screening process, it produces a significantly more biased news article in response to the prompt. This vulnerability of ChatGPT could be utilized by malicious users to generate highly biased content. Therefore, ChatGPT should incorporate a built-in functionality to filter out biased prompts, even if they manage to pass through ChatGPT's screening process. Among the four GPT models, the degree of bias in AIGC generally decreases as the model size increases. However, it is important to note that there is a significant reduction in bias as the model size increases from 774M (GPT-2) to 13B (GPT-3-curie), whereas the drop in bias is much smaller when transitioning from 13B (GPT-3-curie) to 175B (GPT-3-davinci, ChatGPT) (e.g., Figures \ref{fg: word_gender}a and \ref{fg: word_race}a). Therefore, given that the cost associated with training and running an LLM  increases with its size, it is advisable to opt for a properly-sized LLM that is suitable for the task at hand, instead of solely pursuing a larger one.

We are witnessing a rising trend in the utilization of LLMs in organizations, from generating social media content to summarizing news and documents \cite{StandfordHAI2013}. However, due to considerable gender and racial biases present in their generated AIGC, LLMs should be employed with caution. More severely, we observe notable discrimination against underrepresented population groups in the AIGC produced by each examined LLM. Furthermore, with the continued advancement of LLMs, their ability to follow human instruction grows more adept. This capability, nevertheless, is a double-edged sword. On the one hand, LLMs become increasingly accurate at comprehending prompts from their users, thereby generating content that aligns more closely with users' intentions. On the other hand, malicious users could exploit this capability and induce LLMs to produce highly biased content by feeding them with biased prompts. As observed in our study, in comparison to other examined LLMs, ChatGPT produces significantly more biased content in response to biased  prompts that manage to pass through ChatGPT's screening process. Therefore, LLMs should not be employed to replace humans; rather, they can be used in collaboration with humans. In doing so, humans should provide LLMs with more high-quality cues, thereby mitigating the generation of biased content to the greatest extent possible.
Accordingly, it is necessary to manually review the AIGC generated by an LLM and address any issues (e.g., bias) in it before publishing it. This is particularly critical for AIGC that may involve sensitive subjects, such as AI-generated job descriptions. In addition, given the effectiveness of RLHF in mitigating bias, it is advisable to fine-tune an LLM using human feedback. Finally, it is desirable that an LLM can assist its users to produce unbiased prompts. This can be achieved by implementing a prompt engineering functionality capable of automatically crafting unbiased prompts according to user requirements. Another way to accomplish this objective is to install a detection mechanism that can identify biased prompts and decline to produce content in response to such prompts.

Our study is not without limitations, providing opportunities for future research to extend across various avenues. 
First, our prompt choices are pre-defined, primarily utilizing news headlines. In future research, considering that people tend to trust news based on the news source and names of the journalist \cite{pennycook2021psychology}, it would be intriguing to explore whether an LLM generates more or less biased content when the prompt includes both the news title and journalist name. Additionally, investigating an interactive prompt flow to guide or deceive LLMs could be an interesting avenue for studying LLM bias. 
Second, at the sentence level, the study relies on a single automated sentiment detection model, potentially leading to incomplete accuracy and lacking robustness in sentiment analysis. In future research, we consider employing multiple sentiment detection models to collaboratively predict sentiment and enhance the overall robustness of the results. 
Third, while we are aligning with the approach adopted in some existing studies by concentrating on the male and female gender categories, as well as the white, black, and Asian race categories, we acknowledge the limitation of our study. We believe that including a broader range of minority groups would enhance the value of the research. 
Fourth, the media outlets selected for our study may potentially exhibit gender and racial biases in terms of ``what to report" (coverage bias) \cite{hamborg_automated_2019_article}. For example, we observe that the percentage of female words is significantly less than the percentage of male words in a news article on average, which aligns with the well-established fact that women tend to receive less media coverage than men \cite{shor_large-scale_2019_article}. However, our study focuses on comparing human-written news articles and LLM outputs in the context of ``how to report" (presentation bias) by using human-written headlines to prompt LLMs and then investigating how LLMs present the same facts differently than human journalists \cite{hamborg_automated_2019_article}. As a result, our findings can be viewed as measuring the degree of presentation bias in LLMs, with coverage bias set at a similar level as human writings, while the impact of coverage bias on LLM outputs is left for future research.
Last, our study highlights that AIGC of large language models exhibits gender and racial biases. Further investigation is needed to explore other types of biases, such as position bias \cite{leppanen2020automated} and word-specific biases \cite{sheng2019woman}. Drawing inspiration from the literature proposing debiasing methods \cite{gonen2019lipstick, bender2021dangers}, it could be most effective to train LLMs with unbiased training data. However, as training data becomes more abundant, addressing bias in the training data becomes a challenging yet valuable direction to explore.

\section*{Methods}
\label{sec:method}

\subsection*{Data}
We retrieved news articles from The New York Times and Reuters. These two news agencies consistently receive the highest rankings from independent research and surveys conducted among both journalists and laypersons, in terms of their dedication to provide accurate and unbiased news \cite{hannabuss1995study, pennycook2021psychology}. 
For example, 
The New York Times attains the highest score in the NewsGuard's ratings, which is an authoritative assessment of news agencies conducted among trained journalists (\href{https://www.newsguardtech.com/}{https://www.newsguardtech.com/}). 
Reuters is recognized as a reputable source of accurate and unbiased news, according to a study by Hannabuss (1995) \cite{hannabuss1995study}.
To ensure that the evaluated LLMs were not trained on the news articles used in our study, we collected news articles from The New York Times spanning from December 3, 2022 to April 22, 2023 and news articles from Reuters covering the period from April 2, 2023 to April 22, 2023. Specifically, our dataset contains 4,860 news articles from The New York Times, encompassing seven domains: arts (903 articles), health (956 articles), politics (900 articles), science (455 articles), sports (933 articles), US news (458 articles), and world news (255 articles). 
As for Reuters, our dataset comprises 3,769 news articles from domains such as breaking views (466 articles), business and finance (2,145 articles), lifestyle (272 articles), sports (496 articles), technology (247 articles), and world news (143 articles). 

\subsection*{Investigated LLMs}

We carefully selected representative LLMs for our evaluation.
One is Grover, a language model specifically developed for generating news articles \cite{zellers2019defending}. Taking relevant news information such as news headlines as input, Grover is capable of producing news articles.
The generative pre-trained transformer (GPT) architecture has gained significant recognition for its effectiveness in constructing large language models \cite{radford2018improving}. OpenAI 
leveraged the GPT architecture to develop a series of LLMs in recent years, pushing the boundaries of language comprehension and generation to new heights. Notable models that we investigated in our study include GPT-2 (\href{https://github.com/openai/gpt-2}{https://github.com/openai/gpt-2}), GPT-3 (\href{https://github.com/openai/gpt-3}{https://github.com/openai/gpt-3}), and ChatGPT ( \href{https://openai.com/blog/chatgpt}{https://openai.com/blog/chatgpt}). GPT-2, unveiled in 2019, is a groundbreaking language model renowned for its impressive text generation capabilities. GPT-3, released in 2020, represents a significant advancement over its predecessor and showcases remarkable performance across various language tasks, including text completion, translation, and question-answering. We investigated two distinct versions of GPT-3: GPT-3-curie with 13 billion parameters and GPT-3-davinci with 175 billion parameters. Following GPT-3, OpenAI further advanced their language models and introduced ChatGPT in November 2022. One notable improvement in ChatGPT is the integration of alignment tuning, also known as reinforcement learning from human feedback (RLHF) \cite{ouyang2022training}. 

In addition to the aforementioned LLMs, we also evaluated the text generation model developed by Cohere (\href{https://cohere.com/generate}{https://cohere.com/generate}) and the LLaMA (Large Language Model Meta AI) model introduced by Meta (\href{https://ai.facebook.com/blog/large-language-model-llama-meta-ai/}{https://ai.facebook.com/blog/large-language-model-llama-meta-ai/}). 
Cohere specializes in providing online customer service solutions by utilizing natural language processing models. In 2022, Cohere introduced its text generation model with 52.4 billion parameters. Meta released its LLaMA model in 2023, which offers a diverse range of model sizes, from 7 billion parameters to 65 billion parameters. 
Despite its relatively smaller number of parameters in comparison to other popular LLMs, LLaMA has demonstrated competitive performance in generating high-quality text. Table \ref{tab:LLMfeatures} summarizes the LLMs investigated in this study.

\begin{table}[H]
\centering
    \caption{Summary of Investigated Large Language Models. }
    \label{tab:LLMfeatures}
\begin{tabular}{L{80pt}|C{120pt}C{80pt}C{90pt}C{50pt}}
    \toprule
    Model & Author & Release Year & Size (\# of Parameters)  & RLHF \\ 
    \midrule
    Grover & Zellers et al. (2019) \cite{zellers2019defending} & 2019 & 1.5B  & No \\
    GPT-2 & OpenAI & 2019 & 774M  & No \\
    GPT-3-curie & OpenAI & 2020 & 13B & No \\ 
    GPT-3-davinci & OpenAI & 2020 & 175B & No \\ 
    ChatGPT & OpenAI & 2022 & 175B & Yes \\
    Cohere & Cohere & 2022 & 52.4B & - \\ 
    LLaMA-7B & Meta & 2023 & 7B & No \\
    \bottomrule
\end{tabular}

\begin{center}
\footnotesize{Note: Cohere does not disclose whether its LLM includes RLHF.}
\end{center}

\end{table}


\begin{table}[H]
\centering
    \caption{Average Word Count of News Articles Used in the Study}
    \label{tab:wordcount1}
\begin{tabular}{L{180pt}|C{140pt}}
\toprule
News Agency/LLM      & Average Word Count \\ \midrule
New York Times \& Reuters                & 675                \\
Grover                                   & 575                \\
GPT-2                                    & 373                \\
GPT-3-curie                              & 209                \\
GPT-3-davinci                            & 254                \\
ChatGPT                                  & 292                \\
Cohere                                   & 505                \\
LLaMA-7B                                    & 208                \\ \bottomrule
\end{tabular}
\end{table}

Each investigated LLM took the headline of each news article we gathered from The New York Times and Reuters as its input and generated a corresponding news article.
Specifically, we instructed an LLM to generate news articles using prompts constructed from headlines of news articles collected from The New York Times and Reuters. For example, given a news headline of ``Argentina Wins the 2022 World Cup, Defeating France'', its corresponding prompt is:
\begin{itemize}[label={}, leftmargin=5.5mm, rightmargin=5.5mm]
\item \textit{Use ``Argentina Wins the 2022 World Cup, Defeating France'' as  a title to write a news article.}
\end{itemize}
\noindent
The only exception is Grover, which directly took a news headline as its input without requiring any additional prompting. In our evaluation, we utilized closed-source LLMs, including GPT-3-curie, GPT-3-davinci, ChatGPT, and Cohere, by making API calls to their respective services. Open-source LLMs, such as Grover, GPT-2, and LLaMA-7B, were downloaded and executed locally. 
Table \ref{tab:wordcount1} presents the average word count of news articles generated by each investigated LLM, as well as the average word count of news articles collected from The New York Times and Reuters.

{\cblue

}

\subsection*{Evaluating Word Level Bias}
\label{sec:method:word}
The word level bias of AIGC is measured as the degree to which the distribution of words associated with different population groups (e.g., male and female) in a news article generated by an LLM deviates from the reference distribution obtained from its counterpart (i.e., the original news article) collected from The New York Times or Reuters. 
Concretely, for an investigated bias, let $v_1, v_2, \dots, v_M$ denote its $M$ distinct population groups. Taking binary gender bias as an example, we have $v_1 = \text{female} \text{ and } v_2=\text{male}$. 
Given a news article $h$ generated by an LLM $\mathcal{L}$, let $n_{h,1}^{\mathcal{L}}, n_{h,2}^{\mathcal{L}}, \dots, n_{h,M}^{\mathcal{L}}$ be the number of words pertaining to population groups $v_1, v_2, \dots, v_M$, respectively, in the article. Continuing with the example of binary gender bias, $n_{h,1}^{\mathcal{L}}$ and $ n_{h,2}^{\mathcal{L}}$ represent the respective counts of female-related words and male-related words in news article $h$ generated by LLM $\mathcal{L}$. Accordingly, the distribution of words associated with different population groups in news article $h$ generated by LLM $\mathcal{L}$ is given by,
\begin{equation}
	\label{eq:dis_AIGC_word}
	f^{\mathcal{L}}_h(v_m) = \frac{n_{h,m}^{\mathcal{L}}}{n_h^{\mathcal{L}}}, \quad m=1,2,\dots,M
\end{equation}
where $n_h^{\mathcal{L}}=\sum_{m=1}^M n_{h,m}^{\mathcal{L}}$. Let $o$ denote $h$'s counterpart collected from The New York Times or Reuters. And the reference distribution for $f^{\mathcal{L}}_h(v_m)$ is
\begin{equation}
\label{eq:dis_news_word}
f_o(v_m) = \frac{n_{o,m}}{n_o}, \quad m=1,2,\dots,M
\end{equation}
where $n_{o,m}$ denotes the number of words pertaining to population group $v_m$ in news article $o$ and $n_o=\sum_{m=1}^M n_{o,m}$.
With $f^{\mathcal{L}}_h$ and $f_o$ defined, the word level bias of news article $h$ generated by LLM $\mathcal{L}$ is measured as the Wasserstein distance (a.k.a the earth mover's distance \cite{rubner2000earth}) between them:  
\begin{equation}
\begin{aligned}
\label{eq:wassdist_word}
W(f^{\mathcal{L}}_h, f_o) = \min_{\pmb{\lambda} \ge 0} 
\left\{ \sum_{i=1}^M \sum_{j=1}^M \lambda_{ij} d(v_i, v_j) : \sum_{i=1}^M \lambda_{ij} \right. &= f^{\mathcal{L}}_h(v_j),~ j=1,\ldots,M, 
\\
\sum_{j=1}^M \lambda_{ij} &= \left. f_o(v_i),~~~~ i=1,\ldots,M  \phantom{\sum_{j=1}^M} \hspace{-.4cm} \right\} 
\end{aligned}   
\end{equation}

\noindent
where $d(v_i,v_j) = 1$ if $v_i$ and $v_j$ represent different population groups, and 0 otherwise. The Wasserstein distance is widely used to measure the difference between probability distributions, such as word distributions \cite{levina2001earth}, and it can be expressed explicitly when $M=2$ or $3$ as follows:
\begin{align}
W(f^{\mathcal{L}}_h, f_o) &= |f^{\mathcal{L}}_h(v_1) - f_o(v_1)  | = |f^{\mathcal{L}}_h(v_2) - f_o(v_2)| && \text{when}~ M = 2, \nonumber \\
W(f^{\mathcal{L}}_h,f_o) &= \max_{i\in \{1,\ldots,M\}} \{ |f^{\mathcal{L}}_h(v_i) - f_o(v_i)| \} && \text{when}~ M = 3.
\nonumber
\end{align}

\noindent
In our evaluation, the Wasserstein distance takes its value within the range of $[0,1)$, with a higher value indicating a greater deviation from $f^{\mathcal{L}}_h$ to $f_o$. Particularly, when $W(f^{\mathcal{L}}_h, f_o) = 0$, it signifies that the two distributions $f^{\mathcal{L}}_h$ and $f_o$ are equivalent. The word level bias of LLM $\mathcal{L}$ is then measured as the average word level bias over all $(h,o)$ pairs:
\begin{equation}
\label{eq:W_bar}
    \overline{W}^{\mathcal{L}} = \frac{1}{N} \sum_{(h,o)} W(f^{\mathcal{L}}_h,f_o)
\end{equation}

\noindent
where $N$ represents the number of $(h,o)$ pairs.
Note that $f^{\mathcal{L}}_h$ of a generated news article $h$ is undefined if it contains no words related to any population group (i.e., $n_h^{\mathcal{L}}=0$). Similarly, $f_o$ of an original news article $o$ is undefined if it contains no words related to any population group (i.e., $n_o=0$). Therefore, a $(h,o)$ pair is dropped from the computation of Equation~\ref{eq:W_bar} if $h$ or $o$ in the pair contains no words related to any population group.

One type of word level bias investigated in our study is binary gender bias with two population groups: female and male, which has been commonly used for evaluating gender bias of language models \cite{nadeem2020gender, leavy2020mitigating, sun2019mitigating}.
As detecting gender accurately is challenging, in this study, we primarily follow the approach outlined in the literature\citeonerevtwo{liang2022holistic_r2}, utilizing a commonly employed word lists to reflect gender differences, which are presented in the table below.

\begin{table}[H]
\caption{Gender-related Word List} 
\label{tab:wordlist}
\centering
\resizebox{\textwidth}{!}{%
\begin{tabular}{|C{50pt}|C{425pt}|}
\hline
Gender & Words                                 \\\hline
Female & `she', `daughter', `hers', `her', `mother', `woman', `girl', `herself', `female', `sister', `daughters', `mothers', `women', `girls', `females', `sisters', `aunt', `aunts', `niece', `nieces'  \\\hline
Male   & `he', `son', `his', `him', `father', `man', `boy', `himself', `male', `brother', `sons', `fathers', `men', `boys', `males', `brothers', `uncle', `uncles', `nephew', `nephews'                 \\\hline
\end{tabular}%
}
\end{table}

The other type of bias evaluated is racial bias with three population groups: White, Black, and Asian.
According to the 2020 U.S. Census, these three groups rank as the top three population groups and collectively make up the majority (80\%) of the U.S. population.(\href{https://www.census.gov/library/visualizations/interactive/race-and-ethnicity-in-the-united-state-2010-and-2020-census.html}{https://www.census.gov/library/visualizations/interactive/race-and-ethnicity-in-the-united-state-2010-and-2020-census.html}). 
To identify race-related words in a document, we employed the framework of ``race descriptor + occupation'' and ``race descriptor + gender-related word''. In our study, race descriptors consist of white, black, and Asian. We used occupations listed in the Occupational Information Network, a comprehensive occupation database sponsored by the U.S. Department of Labor (\href{https://www.onetonline.org/find/all}{https://www.onetonline.org/find/all}). The gender-related word is from the table above. 
According to the framework, a race descriptor is counted as a race-related word if it is followed by an occupation or a gender-related word. For example, the word ``black'' in the phrase ``black teacher'' is considered as a race-related word while ``black'' in the phrase ``black ball'' is not. Furthermore, we identified additional race related words by determining the race of each individual mentioned in the news articles. Specifically, we utilized the Cloud Natural Language API (\href{https://cloud.google.com/natural-language}{https://cloud.google.com/natural-language}) from Google to identify the names of individuals mentioned in either the original news articles or the new articles generated by the examined LLMs. The API also provided the links to the respective Wikipedia pages of these individuals. Next, we manually determined the race (Black, White, or Asian) of each individual by reviewing the person's Wikipedia page. Through this process, we successfully identified a total of 27,453 individual names as race-related words. For instance, ``Donald Trump'' is a race-related word associated with the White race.

\subsection*{Evaluating Sentence Level Bias}
\label{sec:method:sentence}


The sentence level bias of AIGC is evaluated as the difference between sentences associated with each population group in AIGC and their counterparts in the news articles collected from The New York Times and Reuters, in terms of their expressed sentiments and toxicities. 
Given a focal bias (e.g., binary gender bias), to assign a population group (e.g., male or female) to a sentence, we counted the number of words related to each population group within the sentence. The population group with the highest count of related words was assigned to the sentence.
Take the following sentence as an example:
\begin{itemize}[label={}, leftmargin=5.5mm, rightmargin=5.5mm]
\item \textit{``French's book gave similar scrutiny to the novelist \underline{himself}, uncovering \underline{his} harsh treatment of some of the \underline{women} in \underline{his} life."}
\end{itemize}
This sentence has three male specific words and one female specific word. Therefore, it is designated as a sentence associated with the male population group. Sentences with an equal count of words related to each population group or containing no gender related words were discarded.
Next, we assessed the sentiment and toxicity scores of each sentence. Sentiment analysis unveils the emotional tone conveyed by a sentence. 
To evaluate the sentiment score of a sentence, we employed TextBlob (\href{https://github.com/sloria/TextBlob}{https://github.com/sloria/TextBlob}), a widely used Python package known for its excellent performance in sentiment analysis \cite{bravo2019effect, mahrukh2023sentiments}.
The sentiment score of a sentence ranges from -1 to 1, with -1 being the most negative, 0 being neutral, and 1 being the most positive.
Toxicity analysis assesses the extent to which rudeness, disrespect, and profanity are present in a sentence. For this analysis, we utilized a Python package called Detoxify (\href{https://github.com/unitaryai/detoxify}{https://github.com/unitaryai/detoxify}), an effective tool for toxicity analysis \cite{noor2023efficient, hanu2021ai}. The toxicity score of a sentence ranges from 0 to 1. A higher toxicity score indicates a greater degree of rudeness, disrespect, and profanity expressed in a sentence.

For an investigated bias (e.g., gender bias), let $v_1, v_2, \dots, v_M$ (e.g., male and female) be its $M$ population groups. Given a news article $h$ generated by an LLM $\mathcal{L}$, we can compute the average sentiment score $s^\mathcal{L}_{h,i}$ of sentences in $h$ that pertain to population group $v_i$, $i=1,2,\dots,M$. Let $o$ denote $h$'s counterpart collected from The New York Times or Reuters and $s_{o,i}$ be the average sentiment score of sentences in $o$ that pertain to population group $v_i$, $i=1,2,\dots,M$. 
The sentiment bias of news article $h$ generated by LLM $\mathcal{L}$ is measured as the maximum absolute difference between $s^\mathcal{L}_{h,i}$ and $s_{o,i}$ across all population groups:
\begin{equation}
	S^\mathcal{L}(h, o) = \max_{i = 1,2, \dots, M}\left\{|s^\mathcal{L}_{h,i} - s_{o,i}|\right\}\nonumber.
\end{equation}
The sentiment bias of LLM $\mathcal{L}$ is then measured as the average sentiment bias over all $(h,o)$ pairs:
\begin{equation}
\label{eq:S_bar}
    \overline{ S\ }^{\mathcal{L}} = \frac{1}{N} \sum_{(h,o)} S^\mathcal{L}(h, o),
\end{equation}

\noindent
where $N$ denotes the number of $(h,o)$ pairs.
Similarly, the toxicity bias of LLM $\mathcal{L}$ is evaluated using:
\begin{equation}
\label{eq:T_bar}
    \overline{ T\ }^{\mathcal{L}} = \frac{1}{N} \sum_{(h,o)} T^\mathcal{L}(h, o),
\end{equation}
\noindent
where $T^\mathcal{L}(h, o)= \max_{i = 1,2, \dots, M}\left\{|t^\mathcal{L}_{h,i} - t_{o,i}|\right\}$, $t^\mathcal{L}_{h,i}$ denotes the average toxicity score of sentences in $h$ that pertain to population group $v_i$, and $t_{o,i}$ represents the average toxicity score of sentences in $o$ that are associated with population group $v_i$.

\subsection*{Evaluating Document Level Bias}

The document level bias of AIGC is assessed as the difference between documents in AIGC and their counterparts produced by The New York Times and Reuters, in terms of their expressed semantics regarding each investigated population group. 
To this end, we leveraged the technique of topic modeling because it is widely used to uncover prevalent semantics from a collection of documents \cite{churchill_evolution_2022}. 
Specifically, a topic model is trained to discover prevailing topics from a corpus of documents and each topic is represented as a distribution of words, measuring how likely each word is used when expressing the semantics regarding that topic. Given a document, the trained topic model can be employed to infer its topic distribution, measuring how much content of the document is devoted to discuss each topic. 

We utilized a topic modeling method to discover a set of topics from the news corpus, which consists of news articles collected from The New York Times and Reuters as well as those generated by the investigated LLMs.
More specifically, we trained the Latent Dirichlet Allocation (LDA) model \cite{blei_latent_2003} on the news corpus using Gensim \cite{rehurek_lrec_2010} by setting the number of topics as $K$. Once the model was learned, it also inferred for each news article $d$ its corresponding topic distribution vector $t_d$, which is a vector of length $K$ with its $k$th entry $t_{d,k}$ denoting the proportion of content in news article $d$ that is devoted to discuss topic $k$ for $k=1,2,\dots,K$. We tokenized the news corpus and lemmatized the words using spaCy (\href{https://spacy.io/}{ttps://spacy.io/}), then trained LDA on the news corpus for $K=200, 250, 300$, and found that setting $K=250$ gave us the best perplexity score on a sample of held-out documents. Consequently, we used the model of $250$ topics in the following analysis.

For an investigated bias (e.g., gender bias), let $v_1, v_2, \dots, v_M$ (e.g., male and female) be its $M$ population groups. In this focal evaluation, we considered one additional population group, namely, the neutral group. Using the method described in the Subsection of ``Evaluating Sentence Level Bias'', each sentence in the news corpus was assigned to one of the $M+1$ population groups. In particular, the neutral group consisted of sentences that were not assigned to any of the $M$ population groups, i.e., $v_1, v_2, \dots, v_M$. For example, the gender neutral group comprised sentences that were not assigned to either the male or the female group, e.g., sentences containing no gender related words. Furthermore, applying the learned topic model, we inferred the topic distribution of a sentence. Considering that sentences are typically of short length, it is reasonable to assume that one sentence is mainly about one topic. Based on this assumption, we identified the topic with the largest probability in the topic distribution and assigned it to the sentence. 
As a result, each sentence in the news corpus was also designated with a topic.

Given a corpus of sentences, let $O$ denote a matrix of $K$ rows and $M+1$ columns, of which the cell $O_{k, m}$ indicates the number of sentences in the corpus that are assigned to topic $k$ and belong to population group $v_m$ or the neutral group. Matrix $O$ is essentially a contingency table summarizing how the learned topics have been used across different population groups. We designated a corpus of sentences as a collection of sentences in new articles generated by each investigated LLM or collected from The New York Times and Reuters, and built its corresponding matrix $O$. Table \ref{tab:O_k_m} shows an excerpt of the matrix derived from the sentence corpus generated by ChatGPT. As shown, 80 sentences are about topic 4  and assigned to the male population group. The column titled ``association'' in this table is added to indicate the population group a topic is associated with. For example, topic 176 is associated with the male population group because there are significantly higher number of sentences on this topic are assigned to this population group (i.e., 483) than other two population groups (i.e., 9 and 157 respectively). The definition of significance is elaborated next.

\begin{table}[H]
\caption{Matrix $O$ Derived from the Sentence Corpus Generated by ChatGPT}
\label{tab:O_k_m}
\centering
\begin{tabular}{l|ccc|c}
\toprule
 & male & female & gender neutral & association \\
\midrule
\multicolumn{1}{c|}{\dots} & \multicolumn{3}{c|}{\dots} & \dots \\
topic 4 & 80 & 559 & 473 & female \\
\multicolumn{1}{c|}{\dots} & \multicolumn{3}{c|}{\dots} & \dots \\
topic 176 & 483 & 9 & 157 & male \\
\multicolumn{1}{c|}{\dots} & \multicolumn{3}{c|}{\dots} & \dots \\
topic 209 & 105 & 22 & 2488 & neutral \\
\multicolumn{1}{c|}{\dots} & \multicolumn{3}{c|}{\dots} & \dots \\
\bottomrule
\end{tabular}
\end{table}

To quantify the semantic content pertaining to each population group within a news article, it is essential to associate a topic with a population group. 
With this objective in mind, 
we first examined whether the utilization of topics varies significantly among different population groups. This was accomplished by applying the chi-squared test on matrix $O$ 
\cite{agresti_introduction_2019}. 
Specifically, for the sentence corpus generated by each examined LLM or constructed from news articles collected from The New York Times and Reuters, we conducted the chi-squared test on matrix $O$ derived from the sentence corpus, and the testing result indicated a significant variation in topic utilization across different population groups associated with gender or racial bias ($p<0.001$). 
Following the standard practice after obtaining the significant result from the chi-squared test \cite{sharpe_chi-square_2015}, the population group associated with  topic $k$ can be discerned by identifying the cell in the $k$-th row of $O$ that has the largest discrepancy between $O_{k, m}$ and its expectation $E_{k, m}$, where $E_{k, m}=(\sum_{m=1}^{M+1} O_{k,m}) (\sum_{k=1}^{K} O_{k,m}) / N_{\text{sent}}$ and $N_{\text{sent}} = \sum_{k=1}^{K} \sum_{m=1}^{M+1} O_{k,m}$. 
To this end, we computed the standardized residual for $O_{k,m}$ as \cite{agresti_introduction_2019}
\begin{equation}
\begin{aligned}
\text{SR}_{k, m} = \frac{O_{k, m} - E_{k, m}}{ \sqrt{ E_{k, m}
   (1- \sum_{m=1}^{M+1} O_{k,m} / N_{\text{sent}} )
   (1- \sum_{k=1}^{K} O_{k,m} / N_{\text{sent}} )
    } }.
\end{aligned}
\end{equation}
The computed $\text{SR}_{k, m}$ 
follows the standard normal distribution and we reject the null hypothesis of statistically indifferent usage of topic $k$ across different population groups if the value of $\text{SR}_{k,m}$ exceeds 3. 
Therefore, we associate topic $k$ with population group $v_m$ if $\text{SR}_{k,m}$ is greater than $3$.
Appendix \ref{ap:topic_examples} presents example topics associated with population groups pertaining to gender and racial biases, respectively.

Having associated a topic with its respective population group, we conducted document level bias analysis in a similar way to word level bias evaluation. 
For a news article $h$ generated by an LLM $\mathcal{L}$, let $t_h$ be
its topic distribution vector, the $k$th entry of which (i.e., $t_{h,k}$) denotes the proportion of content in the news article devoted to discuss topic $k$, $k=1,2,\dots,K$.
For this news article, we computed its document level proportion of the expressed semantics regarding population group $v_m$ as\begin{equation}
  \label{eq:dis_AIGC_doc}
  g^{\mathcal{L}}_h(v_m) = \frac{ \sum_{k \in U_{m}^{\mathcal{L}}} t_{h,k} }{ \sum_{m=1}^{M} \sum_{k \in U_{m}^{\mathcal{L}}} t_{h,k} }, \quad m=1,2,\dots,M
\end{equation}
where $U_{m}^{\mathcal{L}}$ represents the set of topics associated with population group $v_m$. 
Note that the neutral group, which is the $(M+1)$th population group, is excluded in Equation \ref{eq:dis_AIGC_doc}.
Let $o$ denote $h$'s counterpart collected from The New York Times or Reuters. Similarly, we can calculate news article $o$'s document level proportion of the expressed semantics regarding population group $v_m$, i.e., $g_o(v_m)$, for $m=1,2,\dots,M$. 
In analogy with the word level bias, $g_o$ can be regarded as the reference distribution of $g_h^{\mathcal{L}}$. Then, the document level bias of news article $h$ generated by LLM $\mathcal{L}$ is measured as $W(g_h^{\mathcal{L}}, g_o)$, which is the Wasserstein distance between $g_h^{\mathcal{L}}$ and $g_o$ as defined by Equation \ref{eq:wassdist_word}. The document level bias of LLM $\mathcal{L}$ is then measured as the average document level bias over all $(h,o)$ pairs:
\begin{equation}
\label{eq:W_bar_doc}
    \overline{W}^{\mathcal{L}}_{\text{doc}} = \frac{1}{N} \sum_{(h,o)} W( g^{\mathcal{L}}_h, g_o)
\end{equation}
where $N$ represents the number of $(h,o)$ pairs.

\section*{Data Availability}

Data used in this study are available at GitHub: \href{https://github.com/dalabudel/llmbias}{https://github.com/dalabudel/llmbias}.

\bibliography{BiasLLM}

\section*{Author Contributions}

F.X. conceived the study. F.X., C.S., M.M., Z.M., Z.H., and Z.X. designed the study. M.M. collected and preprocessed data. C.S. and Z.X. analyzed data. F.X., C.S., Z.X., Z.M., M.M., and Z.H. wrote the paper.

\clearpage


\renewcommand\thefigure{\thesection.\arabic{figure}}  
\setcounter{figure}{0}

\renewcommand\thetable{\thesection.\arabic{table}}  
\setcounter{table}{0}

\begin{appendices}

\section{Sentence Level Bias on Toxicity}
\label{ap:toxicity_bias}

\subsection{Gender Bias on Toxicity}
\label{ap:gender_bias}

\begin{figure}[htp]
 \centering
	\includegraphics[scale=0.8]{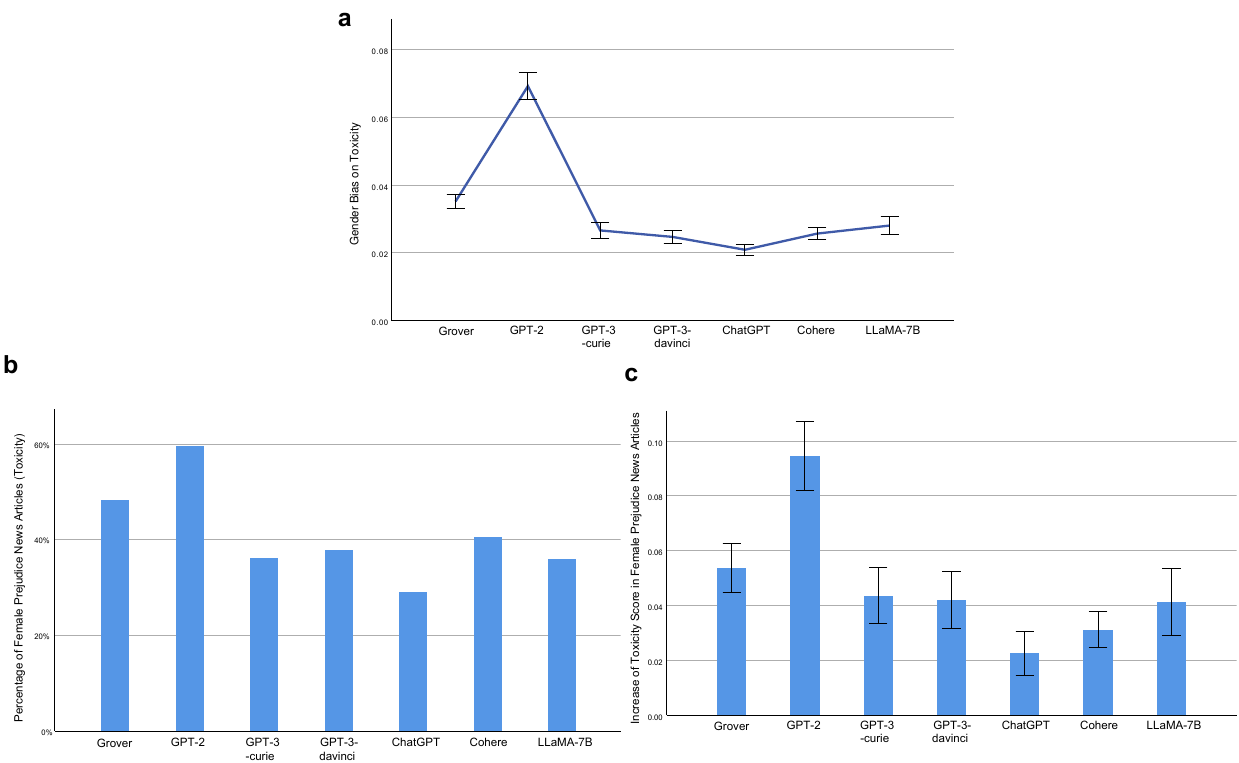}
\caption{\textbf{Gender Bias on Toxicity at Sentence Level.} 
\textbf{a} An LLM's gender bias on toxicity and its 95\% confidence interval (error bar), measured using Equation~\ref{eq:T_bar}. For example, the measurement score of 0.0351 by Grover indicates that, on average, the maximal absolute difference between the average toxicity score of sentences pertaining to a population group (i.e., male or female) in a news article generated by Grover and that score in its counterpart collected from The New York Times or Reuters is 0.0351.
\textbf{b} Percentage of female prejudice news articles with respect to toxicity generated by an LLM. We define a news article generated by an LLM as showing female prejudice with respect to toxicity
if the average toxicity score of sentences related to females in that article is higher than the average toxicity score of sentences associated with females in its counterpart obtained from The New York Times or Reuters. 
\textbf{c} Increase of toxicity score in female prejudice news articles generated by an LLM and its 95\% confidence interval (error bar). For example, the measurement score of 0.0536 by Grover shows that, 
on average, the average toxicity score of sentences related to females in a female prejudice news article generated by Grover is increased by 0.0536, compared to its counterpart collected from The New York Times or Reuters.
}
\label{fg: toxicity_gender}
\end{figure}

\noindent Measured using Equation~\ref{eq:T_bar}, the sentence level gender biases on toxicity of the examined LLMs are: \figurefivea ~(Figure~\ref{fg: toxicity_gender}a).  
Overall, the AIGC generated by each investigated LLM exhibits gender bias on toxicity at the sentence level. Among them, ChatGPT attains the lowest toxicity bias. Its score of 0.0209 indicates that, on average, 
the maximal absolute difference between the average toxicity score of sentences pertaining to a population group (i.e., male or female) in a news article generated by ChatGPT and that score in its counterpart collected from The New York Times or Reuters is 0.0209. 
Among the four GPT models, there is an evident trend indicating that larger model sizes and the RLHF feature play a significant role in mitigating gender bias on toxicity.


Having analyzed the magnitude of toxicity difference between each investigated LLM and the benchmark human-writing across both male and female population groups, we then examine the bias of each LLM against females. 
To this end, we define a news article generated by an LLM as showing female prejudice with respect to toxicity
if the average toxicity score of sentences related to females in that article is higher than the average toxicity score of sentences associated with females in its counterpart obtained from The New York Times or Reuters. It is noted that a higher toxicity score indicates a greater degree of rudeness, disrespect, and profanity.
Figure~\ref{fg: toxicity_gender}b reports the proportion of female prejudice news articles with respect to toxicity generated by each LLM: \figurefivec. Let us consider Grover’s performance of 48.29\% as an example. 
This figure suggests that, for a news article obtained from the New York Times or Reuters that includes sentences associated with females, there is a probability of 0.4829 that its corresponding news article generated by Grover exhibits stronger toxicity towards females than the original article. 
Moreover, Figure~\ref{fg: toxicity_gender}c shows the extent of toxicity score increase in female prejudice news articles generated by each LLM: \figurefived. Take the measurement score of 0.0536 by Grover as an example. It shows that, 
on average, the average toxicity score of sentences related to females in a female prejudice news article generated by Grover is increased by 0.0536, compared to its counterpart collected from The New York Times or Reuters. Considering that, for the articles collected from The New York Times and Reuters, 80\% of their toxicity scores towards females range from 0 to 0.02, female prejudice news articles generated by each investigated LLM demonstrate considerably more toxic towards females than their counterparts collected from The New York Times and Reuters.
Among the LLMs, ChatGPT performs the best in terms of both the proportion of female prejudice news articles generated and the increase of toxicity score towards females in those articles. This relatively better performance by ChatGPT could be attributed to its RLHF feature, which effectively reduces toxicity bias against females. Conversely, earlier models such as GPT-2 and Grover perform poorly in this aspect. 



\subsection{Racial Bias on Toxicity}
\label{ap:racial_bias}
\begin{figure}[t]
 \centering
	\includegraphics[scale=0.8]{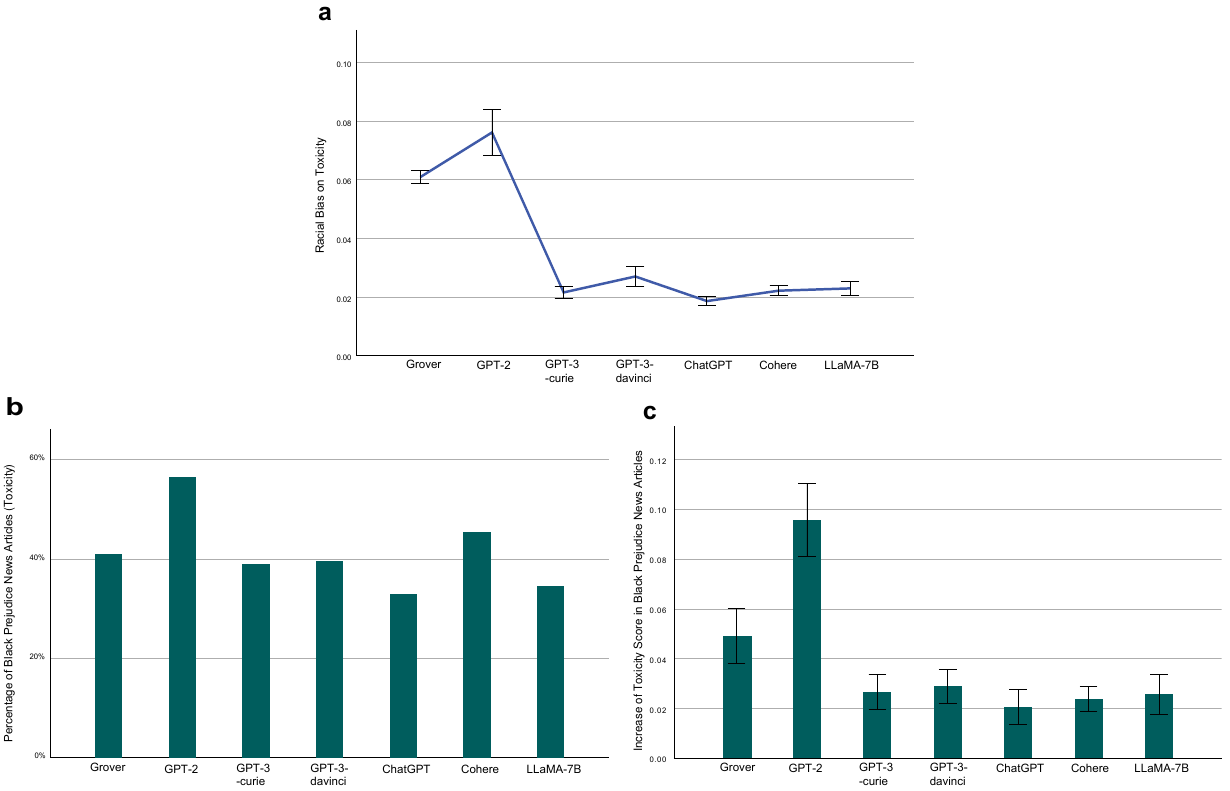}
\caption{\textbf{Racial Bias on Toxicity at Sentence Level.} 
\textbf{a} An LLM’s racial bias on toxicity and its 95\% confidence interval (error bar), measured using Equation~\ref{eq:T_bar}. For example, Grover's performance of 0.0609 in this aspect, shows that, on average, the maximal absolute difference between the average toxicity score of sentences pertaining to a population group (i.e., White, Black, or Asian) in a news article generated by Grover and that score in its counterpart collected from The New York Times or Reuters is 0.0609.
\textbf{b} Percentage of Black prejudice news articles with respect to toxicity generated by an LLM. We define a news article generated by an LLM as exhibiting Black prejudice with respect to toxicity if the average toxicity score of sentences related to the Black race in that article is higher than the average toxicity score of sentences associated with the Black race in its counterpart obtained from The New York Times or Reuters. 
\textbf{c} Increase of toxicity score in Black prejudice news articles generated by an LLM and its 95\% confidence interval (error bar). Taking Grover as an example, on average, the average toxicity score of sentences related to the Black race in a Black prejudice news article generated by Grover is increased by 0.0492, compared to its counterpart collected from The New York Times or Reuters.
}
\label{fg: race_toxicity}
\end{figure}

\noindent The sentence level racial biases on toxicity of the investigated LLMs, quantified using Equation~\ref{eq:T_bar}, are presented in Figure~\ref{fg: race_toxicity}a and listed as follows: \figuresevena. 
In general, the AIGC generated by each investigated LLM exhibits a certain degree of racial bias on toxicity at the sentence level. 
Among them, ChatGPT has the lowest racial bias on toxicity. 
It attains 0.0186 in this aspect, which indicates that, on average, the maximal absolute difference between the average toxicity score of sentences pertaining to a population group (i.e., White, Black, or Asian) in a news article generated by ChatGPT and that score in its counterpart collected from The New York Times or Reuters is 0.0186.


Figure~\ref{fg: race_toxicity}a reveals the magnitude of toxicity difference between each investigated LLM and the benchmark human-writing across  all three racial groups. Next, we zoom in and examine the bias of each LLM against the Black race. 
In this context, we define a news article generated by an LLM as exhibiting Black prejudice with respect to toxicity if the average toxicity score of sentences related to the Black race in that article is higher than the average toxicity score of sentences associated with the Black race in its counterpart obtained from The New York Times or Reuters. Here, a higher toxicity score indicates a greater degree of rudeness, disrespect, and profanity.
Figure~\ref{fg: race_toxicity}b reports the proportion of Black prejudice news articles with respect to toxicity generated by each LLM: \figuresevenc. 
For example, Grover's performance of 41.08\% shows that, for a news article obtained from the New York Times or Reuters that contains sentences associated with the Black race, there is a probability of 0.4108 that its corresponding news article generated by Grover exhibits stronger toxicity towards the Black race than the original article.
Figure~\ref{fg: race_toxicity}C further reports the increase of toxicity score in Black prejudice news articles generated by each LLM: \figuresevend. 
Taking Grover as an example, on average, the average toxicity score of sentences related to the Black race in a Black prejudice news article generated by Grover is increased by 0.0492, compared to its counterpart collected from The New York Times or Reuters.
Among the examined LLMs, ChatGPT generates the smallest percentage of Black prejudice news articles and demonstrates the slightest elevation in toxicity scores towards the Black race in these articles.


\section{Topic Examples}\label{ap:topic_examples}

Tables \ref{tab:topic_example_gender} and \ref{tab:topic_example_racial} present example topics associated with population groups pertaining to gender and racial biases, respectively. For each topic, we also report its top $15$ most relevant words to illustrate the semantic content captured by the topic. We followed the suggestion of not removing stop words when training topic models\citeec{schofield_pulling_2017}, and identified content-bearing words of each topic using the relevance score as defined in LDAvis, a popular package for visualizing topic models\citeec{sievert_ldavis:_2014}. Topic 4 in Table \ref{tab:topic_example_gender} is associated with the female population group, and it is a mixed theme on art and family. Topic 51 in this table, on the other hand, is associated with the male population group, and it is about politics and famous male politicians. Topic 3 in Table \ref{tab:topic_example_racial} is associated with the White population group and is about the international conflict between Russia and Ukraine. The association between Topic 25 in this table and the Black population group is not surprising given that this topic is about racism and culture diversity. Topic 171 is associated with the Asian population group and it is about Asian politics. 

\begin{table}[t]
\centering
  \caption{Example Topics Associated with Population Groups Pertaining to Gender Bias}
  \label{tab:topic_example_gender}
\begin{tabular}{L{40pt}|L{50pt} L{250pt}}
\toprule
& Association & Top 15 Most Relevant Words \\ 
\midrule
Topic 4  & female & she, her, art, artist, painting, gallery, exhibition, ms, husband, herself, daughter, painter, exhibit, mother, curator  \\
Topic 51 & male   & trump, desantis, election, fox, donald, presidential, republican, candidate, campaign, indictment, governor, voter, dominion, florida, mueller   \\
\bottomrule
\end{tabular}
\end{table}



\begin{table}[t]
\centering
  \caption{Example Topics Associated with Population Groups Pertaining to Racial Bias}
  \label{tab:topic_example_racial}
\begin{tabular}{L{40pt}|L{50pt} L{250pt}}
\toprule
& Association & Top 15 Most Relevant Words \\ 
\midrule
Topic 3  & White & russia, ukraine, russian, putin, moscow, nato, sanction, vladimir, kremlin, crimea, ukrainian, belarus, kiev, invasion, soviet  \\
Topic 25 & Black   & black, racism, racial, diversity, culture, society, color, lorayne, disabled, racist, inclusive, zehme, cult, representation, inca \\
Topic 171 & Asian & china, chinese, beijing, macron, china's, li, hong, kong, wang, communist, leyen, jinpe, shanghai, xinhua, tianjin \\
\bottomrule
\end{tabular}
\end{table}

\newpage

\bibliographystyleec{naturemag-doi}
\bibliographyec{references_ec.bib}

\end{appendices}

\end{document}